\documentclass{ecai}

\makeatletter
\g@addto@macro{\UrlBreaks}{\do\-}
\makeatother

\makeatletter
\newcommand{\pdfulink}[1]{%
  \begingroup
    \edef\pdful@uri{\pdfescapestring{#1}}%
    \pdfstartlink attr{/Border[0 0 0]}%
      user{/Subtype/Link /A<< /S/URI /URI(\pdful@uri)>>}%
      \url{#1}
    \pdfendlink
  \endgroup
}
\makeatother

\usepackage{latexsym}
\usepackage{soul}
\usepackage[utf8]{inputenc}
\usepackage{graphicx}
\usepackage{amsmath}
\usepackage{amsthm}
\usepackage{thm-restate}
\usepackage{booktabs}
\usepackage{paralist} %
\usepackage[inline]{enumitem}
\usepackage{algorithmic}
\usepackage[switch]{lineno}
\usepackage[ruled,noend,linesnumbered]{algorithm2e}
\usepackage{tikz}
\usepackage{dsfont}

\usepackage{subfigure}

\usetikzlibrary{decorations.pathreplacing,calligraphy,patterns}
\usepackage{tikz}
\usetikzlibrary{positioning}
\usetikzlibrary{backgrounds}
\usetikzlibrary{fit}
\usetikzlibrary{automata}
\usetikzlibrary{shapes.geometric}
\usetikzlibrary{decorations.pathreplacing,calligraphy}

\newtheorem{definition}{Definition}

\usepackage{amssymb}
\usepackage{thmtools}
\usepackage{mathtools}
\usepackage[capitalise]{cleveref}
\usepackage[all]{foreign}
\usepackage{forest}
\usepackage{multicol}
\usepackage{lipsum}
\usepackage[flushleft]{threeparttable}
\usepackage{multirow}

\usepackage{packages/commands} %
\usepackage{packages/graphics} %
\usepackage{packages/logic}    %
\usepackage{packages/symbols}  %

\usepackage{tabularx}
\usepackage{colortbl}

\newcommand\extrafootertext[1]{%
    \bgroup
    \renewcommand\thefootnote{\fnsymbol{footnote}}%
    \renewcommand\thempfootnote{\fnsymbol{mpfootnote}}%
    \footnotetext[0]{#1}%
    \egroup
}

\newcommand{\fractgood}{\ensuremath{\mathit{fract_{good}}}\xspace}
\newcommand{\quality}{\ensuremath{\mathit{q}}\xspace}
\newcommand{\rotation}{\ensuremath{\mathit{r_{interval}}}\xspace}
\newcommand{\maxgen}{\ensuremath{\mathit{max_{gen}}}\xspace}
\newcommand{\patience}{\ensuremath{\mathit{patience}}\xspace}
\newcommand{\minacc}{\ensuremath{\mathit{min_{acc}}}\xspace}
\newcommand{\maxfar}{\ensuremath{\mathit{max_{far}}}\xspace}
\newcommand{\popsize}{\ensuremath{\mathit{pop_{size}}}\xspace}
\newcommand{\mutprob}{\ensuremath{\mathit{mut_{prob}}}\xspace}
\newcommand{\crossprob}{\ensuremath{\mathit{cross_{prob}}}\xspace}
\newcommand{\epoch}{\ensuremath{\mathit{e}}\xspace}
\newcommand{\batch}{\ensuremath{\mathit{b}}\xspace}
\newcommand{\kopt}{\ensuremath{k_{opt}}\xspace}

\newenvironment{claim}[1]{\par\noindent\underline{Claim:}\space#1}{}
\newenvironment{claimproof}[1]{\par\noindent\underline{Proof:}\space#1}{\hfill $\blacksquare$}

\newcommand{\BibTeX}{B\kern-.05em{\sc i\kern-.025em b}\kern-.08em\TeX}
\newcommand{\stdf}[1]{\mbox{ (#1)}}

\begin{document}

\begin{frontmatter}

\paperid{0414}

\title{
Interpretable Early Failure Detection via \\Machine Learning and Trace Checking-based Monitoring
}

\author[A]{\fnms{Andrea}~\snm{Brunello}\orcid{0000-0003-2063-218X}%
}
\author[A]{\fnms{Luca}~\snm{Geatti}\orcid{0000-0002-7125-787X}} 
\author[A]{\fnms{Angelo}~\snm{Montanari}\orcid{0000-0002-4322-769X}}
\author[A]{\fnms{Nicola}~\snm{Saccomanno}\orcid{0000-0001-5916-3195}\thanks{Corresponding Author. Email: nicola.saccomanno@uniud.it. \\ This work is the full version (appendix and code) of \cite{ecai2025}. Please, cite \cite{ecai2025} when referring to this work.}} 

\address[A]{University of Udine, Italy}

\begin{abstract}
Monitoring is a runtime verification technique that allows one to check whether an ongoing computation of a system (partial trace) satisfies a given formula. It does not need a complete model of the system, %
but it typically requires the construction of a deterministic automaton doubly exponential in the size of the formula (in the worst case), which limits its practicality. 
In this paper, we show that, when considering finite, discrete traces, monitoring of pure past (co)safety fragments of Signal Temporal Logic (\STL) can be reduced to trace checking, that is, evaluation of a formula over a %
trace, that can be performed in time polynomial in the size of the formula and the length of the trace. By exploiting such a result, we develop a GPU-accelerated framework for interpretable early failure detection based on vectorized trace checking, that employs genetic programming to learn temporal properties from historical trace data. The framework shows a 2–10\% net improvement in key performance metrics compared to the state-of-the-art methods.

\end{abstract}

\end{frontmatter}

\section{Introduction}
\label{sec:intro}

System \textit{failures} are undesirable terminating events that one wants to prevent by
detecting anticipatory signals as much in advance as possible.
In this paper, we present a GPU-enabled framework that relies on faulty and faultless labeled historical data to extract new, relevant, and interpretable properties, encoded in \emph{Signal Temporal Logic} (\STL)~\cite{formats/MalerN04}, to 
perform early failure detection with formal guarantees. This stems from a seamless integration of \textit{monitoring} and \emph{machine learning}, with
GPU acceleration made possible by a novel theoretical insight that establishes a connection between \textit{monitoring} and \textit{trace checking}.

Monitoring is a runtime verification technique to analyze complex systems~\cite{leucker2009brief}. It involves generating a \emph{monitor} that runs alongside the system under analysis, producing a positive (resp., negative) verdict as soon as it can establish that all possible continuations of the current execution conform to (resp., deviate from) the intended behavior. Monitor verdicts are thus always \emph{irrevocable}.
Behaviors to be monitored are usually specified in a temporal logic, like, e.g., \emph{Linear Temporal Logic} (\LTL)~\cite{pnueli1977temporal} or \STL,
as in our case~\cite{formats/MalerN04}.

Regardless of the adopted temporal logic, monitoring faces a key challenge that limits its broad applicability: the complexity of modern systems often makes it impractical, if not impossible, for domain experts to specify in advance all the properties to be monitored. Such an issue is exacerbated by the fact that not all temporal formulas are \emph{monitorable}, that is, some of them cannot produce an irrevocable verdict on the basis of a finite number of observations~\cite{bauer2011runtime}. In addition, checking monitorability of a formula is often a costly process %
\cite{DBLP:conf/rv/HavelundP23}. 
For the above reasons, previous contributions, like,  e.g.,~\cite{brunello2019synthesis,brunello2024learning,DBLP:conf/birthday/Neider025,DBLP:conf/cav/ValizadehFB24}, proposed methods to learn relevant properties from historical data, two of them in the specific context of monitoring and early failure detection~\cite{brunello2019synthesis,brunello2024learning}.
These approaches achieved promising results in terms of accuracy and interpretability. 
Yet, when considering large amounts of data, so as to better model the monitored system, 
they showed important scalability limitations (learning properties from data is \NP-HARD for \LTL \cite{DBLP:conf/icgi/FijalkowL21}).
In an effort to expedite learning, Brunello et al.\ proposed to generate only properties within the \emph{pure past safety and cosafety fragments} of \STL (resp., \GppSTL and \FppSTL), showing that all formulas in these syntactic fragments are \emph{monitorable} and (in the case of discrete traces) expressively \textit{complete} w.r.t. all \STL (co)safety languages \cite{brunello2024learning}. Although this reduces the search space and the need to check monitorability of the learned formulas, it is still insufficient. 

In this paper, we focus on accelerating the exploration of the search space of candidate formulas, investigating the conditions under which monitoring can be reduced to another equivalent, but computationally simpler problem, while leveraging GPU parallelization.

Similar to~\cite{brunello2024learning}, we restrict the formula scope to the \GppSTL and \FppSTL fragments, and we show how monitoring, when considering finite, discrete traces, can be effectively reduced to \emph{trace checking}, i.e., verifying whether the current execution trace satisfies the formula. This can be done in polynomial time with respect to the formula size and trace length and, crucially, in a vectorized manner.

On the basis of the above theoretical result, we develop a framework that performs early, interpretable failure detection by automatically learning formulas in \GppSTL (or \FppSTL) only. This is done by means of a grammar-based multi-objective evolutionary algorithm (EA) in which, to efficiently evaluate the quality of candidate formulas, we implemented a GPU-based trace checking module that vectorizes the evaluation of multiple formulas over multiple traces. 
This dramatically accelerates the evaluation of the formulas that are generated during the evolution,  
leading to a more comprehensive and exhaustive exploration of the (infinite) space of candidate formulas, and to a net 2-10\% improvement over state-of-the-art methods across several key metrics and public datasets.

\section{Background}
\label{sec:back}

In this section, we provide  background knowledge on \STL over finite traces,
safety and cosafety fragments, and monitoring.

\subsection{Signal Temporal Logic over finite traces}
From now on, let $\bar{x} = \seq{x_1,\dots,x_n}$, for some $n\ge 0$, be
a sequence of variables each with domain $\R$. The syntax of
\emph{Signal Temporal Logic} is defined as follows~\cite{formats/MalerN04} as follows.

\begin{definition}[Signal Temporal Logic]
\label{def:STL:syntax}
  Formulas $\phi$ of \emph{Signal Temporal Logic} (\STL) are inductively
  defined as follows:
  \begin{align*}
    \phi \coloneqq 
      \top                      \choice
      \mathcal{A}               \choice
      \ltl{! \phi}              \choice 
      \ltl{\phi || \phi}        \choice
      \ltl{X \phi}              \choice
      \ltl{\phi U_I \phi}       \choice
      \ltl{Y \phi}              \choice
      \ltl{\phi S_I \phi}
  \end{align*}
  where $\mathcal{A} = \set{f_z(\bar x) \ge c_z \suchthat z\ge 0}$ such
  that, for each $z\ge 0$, $f_z : \R^n \to \R$ is a linear arithmetic
  function and $c_z \in \R$. In addition, we require $I$ to be an interval
  of the form $[a,b]$ or $[a,\infty)$, where $a,b \in \N$, with $a \le b$,
  are represented in \emph{binary} notation.
\end{definition}

Formulas of the form $f_z(\bar x) \ge c_z$ are called \emph{atomic formulas}.
Modalities $\ltl{U_I}$ and $\ltl{S_I}$ are called \emph{until} and
\emph{since}, respectively. We define the standard shortcut operators as
follows:
\begin{itemize*}%
  \item \emph{weak tomorrow}: $\ltl{wX \phi \coloneqq ! X ! \phi}$;
  \item \emph{eventually}: $\ltl{F_I \phi \coloneqq \true U_I \phi}$;
  \item \emph{globally}: $\ltl{G_I \phi \coloneqq ! F_I ! \phi}$;
  \item \emph{release}: $\ltl{\phi_1 R_I \phi_2 \coloneqq !(!\phi_1 U_I
    !\phi_2)}$;
  \item \emph{weak yesterday}: $\ltl{wY \phi \coloneqq ! Y ! \phi}$;
  \item \emph{once}: $\ltl{O_I \phi \coloneqq \true S_I \phi}$;
  \item \emph{historically}: $\ltl{H_I \phi \coloneqq ! O_I ! \phi}$;
  \item \emph{triggers}: $\ltl{\phi_1 T_I \phi_2 \coloneqq !(!\phi_1 S_I
    !\phi_2)}$.
\end{itemize*}

For $I = [0,\infty)$, we define $\ltl{U}_I$ (respectively, $\ltl{S_I}$) as
\emph{unbounded}, and denote them simply as $\ltl{U}$ (respectively, $\ltl{S}$);
otherwise, they are considered \emph{bounded}. This convention similarly applies
to the other operators. Modalities $\ltl{X}$, $\ltl{U}_I$, and all their derived
shorthand notations are referred to as \emph{future modalities}, whereas
$\ltl{Y}$, $\ltl{S}_I$, and their corresponding shorthand notations are referred
to as \emph{past modalities}. The set of all Signal Temporal Logic
formulas is denoted by \STL.
The \emph{pure past fragment of} \STL, denoted by \ppSTL, is the set of
\STL formulas devoid of future operators.
The \emph{size of $\phi$}, denoted by $|\phi|$, is defined as the number of
symbols in $\phi$.

A \emph{state} $s \in \R^n$ represents the $n$ values $d_1, \dots, d_n$
associated with the variables $x_1, \dots, x_n$, respectively.
A \emph{trace} $\sigma$ is a \emph{finite} (and nonempty)
sequence of such states.  
From now on, we denote by $(\R^n)^+$ the set of all
finite and nonempty traces. Moreover, we define $(\R^n)^* \coloneqq
(\R^n)^+ \cup \epsilon$, where $\epsilon$ is the \emph{empty trace}.
We typically represent $\sigma$ as the sequence
$\langle \sigma_0, \sigma_1, \dots \rangle$. The length of $\sigma$,
denoted $|\sigma|$, corresponds to the number of states in $\sigma$. For
any $0 \leq i < |\sigma|$, $\sigma_{[0, i]}$ denotes the prefix of $\sigma$
up to position $i$. Additionally, for any $\sigma,\sigma' \in (\R^n)^*$, the concatenation of $\sigma'$ to $\sigma$ is denoted by
$\sigma \cdot \sigma'$.  A \emph{language} $\lang$ is a set of traces.

The following sections recall the standard notions of qualitative and
quantitative semantics for \STL. The first %
determines
whether a signal satisfies a given formula, while the second %
assigns a real-valued measure reflecting the degree of
satisfaction or violation. These definitions provide the foundation for our
theoretical contributions in~\cref{sec:mon2trcheck} and %
performance
improvements discussed in~\cref{sec:framework,sec:exper}. Notably,
the quantitative semantics will be used to formulate the numerical fitness
function in the evolutionary algorithm.

\paragraph{Quantitative semantics of \STL over finite traces}
Let $\phi$ be an \STL formula with variables $\bar{x}
= \seq{x_1,\dots,x_n}$, let $\sigma \in (\R^n)^+$ be a trace, and let $0\le
i < |\sigma|$ be a position. We define the \emph{robustness of $\phi$ over
$\sigma$ at position $i$}, denoted as $\rho(\phi,\sigma,i)$, inductively as
follows:
\begin{itemize}
  \item $\rho(\top,\sigma,i) = +\infty$;
  \item $\rho(f(\bar x) \ge c,\sigma,i) = f(\bar x) - c$; 
  \item $\rho(\lnot \phi,\sigma,i) = - \rho(\phi,\sigma,i)$;
  \item $\rho(\phi_1 \lor \phi_2,\sigma,i)
    = \max\{\rho(\phi_1,\sigma,i),\rho(\phi_2,\sigma,i)\}$;
  \item $\rho(\ltl{X \phi_1},\sigma,i) = 
    \begin{cases}
      \rho(\phi_1,\sigma,i+1) &  i<|\sigma| -1 \\
      -\infty & \mbox{otherwise}
    \end{cases}$
  \item $\rho(\ltl{\phi_1 U_{[a,b]} \phi_2},\sigma,i) = \\
    \begin{cases}
      \displaystyle\max_{j\in[i+a,i+b]} \min\set{\rho(\phi_2,\sigma,j), \displaystyle\min_{k\in[i,j)}
        \rho(\phi_1,\sigma,k)}&  i+b<|\sigma| \\
      -\infty & \mbox{otherwise}
    \end{cases}$
  \item $\rho(\ltl{Y \phi_1},\sigma,i) = 
    \begin{cases}
      \rho(\phi_1,\sigma,i-1) & \mbox{if } i>0 \\
      -\infty & \mbox{otherwise}
    \end{cases}$
  \item $\rho(\ltl{\phi_1 S_{[a,b]} \phi_2},\sigma,i) = \\
    \begin{cases}
      \displaystyle\max_{j\in[i-b,i-a]} \min\set{\rho(\phi_2,\sigma,j), \displaystyle\min_{k\in (j,i]}
        \rho(\phi_1,\sigma,k)}& i-b\ge 0 \\
      -\infty & \mbox{otherwise}
    \end{cases}$
\end{itemize}

\paragraph{Qualitative semantics of \STL over finite traces}
Given a \STL formula $\phi$ with variables $\bar{x}
= \seq{x_1,\dots,x_n}$, a trace $\sigma \in (\R^n)^+$, and a position $i$,
with $0\le i<|\sigma|$, the Boolean, qualitative semantics of \STL is
defined according to $\rho(\phi,\sigma,i)$ sign. Formally, we say
that \emph{$\sigma$ satisfies $\phi$ at position $i$}, denoted by
$\sigma,i\models\phi$, iff the following holds:
\begin{itemize} 
    \item  $\sigma,i \models \phi$ iff $\rho(\phi,\sigma,i) \geq 0$ 
\end{itemize}

Formulas of \STL are interpreted at the \emph{first} position of a trace
$\sigma$: we say that \emph{$\sigma$ is a model of the \STL formula $\phi$}
(written $\sigma \models \phi$) iff $\sigma,0 \models \phi$. The language
of a \STL formula $\phi$, denoted with $\lang(\phi)$, is the set of its
models, and is formally defined as follows.

\begin{definition}[Language of an \STL formula]
\label{def:STL:omega:language}
  Given an \STL formula $\phi$ with variables $x_1,\dots,x_n$, \emph{the
  language of $\phi$}, denoted by $\lang(\phi)$, is the language
  $\set{\sigma \in (\R^n)^+ \suchthat \sigma \models \phi}$.
\end{definition}

For every formula $\phi$ of \ppSTL, we interpret $\phi$ at the \emph{last}
time point of a finite trace. The language of $\phi$ is thus defined as
follows:
\begin{align*}
  \lang(\phi) = \set{\sigma \in (\R^n)^+ \suchthat
  \sigma,|\sigma|-1 \models \phi}
\end{align*}

\subsection{Safety and cosafety fragments}
Safety and cosafety properties play a central role in the verification and
the monitoring of critical cyber-physical systems.  %
Safety
properties express the fact that ``\emph{something bad never happens}'',
such as a sensor exceeding a specified temperature threshold.  On the
contrary, cosafety properties express the fact that ``\emph{something good
will eventually happen}'' like, for instance, termination of a program.

We specify the desired properties (or those to be avoided) of a system
using \STL formulas, particularly through the languages they recognize.
The definition of safety and cosafety language~\cite{brunello2024learning}
is reported here below.

\begin{definition}[Safety and cosafety languages]
\label{def:rt:mv:cosafety:lang}
\label{def:rt:mv:safety:lang}
  Let $\lang \subseteq (\R^n)^+$ be a language. We say that $\lang$ is
  a \emph{cosafety (resp. \emph{safety}) language} iff, for all $\sigma \in (\R^n)^+$, if $\sigma
  \in \lang$ (resp. $\sigma
  \not\in \lang$), then there exists $0\le i< |\sigma|$ such that
  $\sigma_{[0,i]} \cdot \sigma' \in \lang$ (resp. $\sigma_{[0,i]} \cdot \sigma' \not\in \lang$), for all $\sigma' \in (\R^n)^*$.
  We say that $\sigma_{[0,i]}$ is a \emph{good prefix} (resp. \emph{bad prefix}) of $\lang$. 
\end{definition}

Intuitively, in cosafety (resp., safety) languages, a finite prefix of
a trace is sufficient to determine whether all its extensions belong
(resp., do not belong) to the language. As we will see in the next section,
this property is particularly useful for the monitoring task.

It holds that $\lang$ is a safety language iff $\overline{\lang}$ is
a cosafety language.  We say that a formula $\phi$ is safety (resp.,
cosafety) iff $\lang(\phi)$ is a safety (resp., cosafety) language.
A syntactic fragment\footnotemark of \STL is a \emph{safety} (resp.,
a \emph{cosafety}) \emph{fragment} iff all formulas $\phi$ definable in the
fragment are safety (resp., cosafety). In~\cite{brunello2024learning},
a safety and a cosafety fragment of \STL are presented. We report their
definition here below.

\footnotetext{%
  A syntactic fragment of \STL is a logic obtained from \STL by forbidding
  the use of some (tipically temporal) operators.
}

\begin{definition}[Safety and cosafety syntactic fragments of \STL]
\label{def:STL:canonical}
  We define the safety (resp., cosafety) syntactic fragment of \STL,
  denoted by \GppSTL (resp., \FppSTL), as the set of \STL formulas of the
  form $\ltl{G}(\psi)$ (resp., $\ltl{F}(\psi)$), where $\psi$ is a formula
  of \ppSTL.
\end{definition}

In~\cite{brunello2024learning}, it is proved that for every formula $\phi$
of \GppSTL (resp., \FppSTL), $\lang(\phi)$ is a safety (resp., cosafety)
language. In addition, it is shown that, when interpreted over \emph{discrete time}, the fragments \GppSTL and \FppSTL are expressively
\emph{complete}, that is, every safety (resp., cosafety) multi-valued
language that can be defined by an \STL formula is definable also in
\GppSTL (resp., in \FppSTL).

\subsection{Monitoring}
\label{sub:monitoring}
Monitoring is a lightweight runtime verification
technique~\cite{leucker2009brief} which involves generating a monitor to
analyze a system's execution either online (at runtime) or offline (e.g., by
processing the system's logs). The monitor outputs either an inconclusive
result (denoted by \textit{?}) or a definitive verdict: a violation (resp.,
satisfaction) if all possible continuations of the observed execution are
bad (resp., good).

\begin{definition}
We define a monitor for a language
$\lang\subseteq (\R^n)^+$ as a function $\mon_{\lang} : (\R^n)^+ \to
\set{\top,\bot,\textit{?}}$ such that, for all $\sigma \in (\R^n)^+$,
\begin{align*}
  \mon_{\lang}(\sigma) \coloneqq
  \begin{cases}
    \top \mbox{ iff } \forall \sigma' \in (\R^n)^* : \sigma\cdot\sigma' \in \lang \\
    \bot \mbox{ iff } \forall \sigma' \in (\R^n)^* : \sigma\cdot\sigma' \not\in \lang \\
    \textit{?} \mbox{ otherwise}
  \end{cases}
\end{align*}
\end{definition}

Given an \STL formula $\phi$, we will denote with $\mon_\phi$ the monitor
$\mon_{\lang(\phi)}$.
When the monitor returns $\top$, we know that all continuations are good
with respect to the property $\phi$: this is the case of cosafety
languages, where $\phi$ expresses the achievement of a goal.  If the
monitor returns $\bot$, we know that there has been an irremediable
violation of $\phi$ (this is the case of safety property).

Not all \STL formulas are monitorable. For example, the \STL formula
$\ltl{F G(x> 0)}$, stating that there exists a position in the future after
which, for all future time points, variable $x$ stabilizes to a strictly
positive value, is \emph{not} monitorable.
However, all cosafety (resp., safety) properties are \emph{positively
monitorable} (resp., \emph{negatively monitorable}) in the sense that, for
every trace $\sigma$, there exists at least one continuation $\sigma'$ such
that $\mon_\phi(\sigma \cdot \sigma') = \top$ (resp., $\mon_\phi(\sigma
\cdot \sigma') = \bot$).

The classical approach to
monitoring~\cite{bauer2011runtime,leucker2009brief} involves constructing,
from a temporal formula $\phi$, a deterministic automaton for the good
prefixes of $\phi$ and one for the bad prefixes of $\phi$. The monitor is
essentially composed of these two automata: as it reads the input trace
(i.e., the system execution), the monitor produces the output $\top$ (if the
automaton for the good prefixes has reached one of its final states) or
$\bot$ (if the automaton for the bad prefixes has reached one of its final
states). In all other cases, the output is \textit{?}. A major limitation
of this approach is that, for \STL formulas (as well as for formulas in
Linear Temporal Logic), the deterministic automata for the good and bad
prefixes have doubly-exponential size in
$|\phi|$~\cite{kupferman2001model}.

\section{\STL monitoring by trace checking}
\label{sec:mon2trcheck}

In this section, we show that for the safety and cosafety fragments of \STL
(\cref{def:STL:canonical}), the monitoring problem can be solved without
resorting to deterministic automata. Specifically, we prove that for
a formula $\phi$ in \GppSTL (resp., %
\FppSTL), monitoring reduces
to the \emph{trace checking} problem, that is, determining whether
a given trace $\sigma$ violates $\phi$ (resp., satisfies $\phi$), i.e.,
$\sigma \not\models \phi$ (resp., $\sigma \models \phi$).
Crucially, %
this allows for the
vectorization of formula computation over traces (\cref{sub:stl_vect}). 
We now formally define the \emph{trace checking} problem. %

\begin{definition}[Trace checking problem]
\label{def:evaluation}
  Let $\sigma \in (\R^n)^+$ be a trace and $\phi$ be a \STL formula
  with variables $\bar{x} = \seq{x_1,\dots,x_n}$. The \emph{trace checking
  problem} is the problem of establishing whether $\sigma \models \phi$.
\end{definition}

Notably, given a trace $\sigma$ and a formula $\phi$ of \STL, the problem
of deciding whether $\sigma \models \phi$ holds can be solved in
$\O(|\sigma| \cdot |\phi|)$ time. This result parallels the case for Linear
Temporal Logic~\cite{kupferman2001model,sistla1985complexity}, with the key
difference being the use of linear real arithmetic constraints in place of
propositional atoms.

The following theorem shows that for any formula $\phi \in \GppSTL$ (resp.,
$\phi \in \FppSTL$) and any trace $\sigma$, determining whether
$\mon_\phi(\sigma) = \bot$ (resp., $\mon_\phi(\sigma) = \top$) is
equivalent to checking whether $\sigma \not\models \phi$ (resp., $\sigma
\models \phi$). This allows us to avoid constructing any form of automaton
for $\mon_\phi$, thus avoiding in the first place the doubly-exponential
blowup of the classical approach to monitoring.

\begin{restatable}{theorem}{montc}
\label{th:mon2tc}
  For all formulas $\phi$ of \FppSTL (resp., of \GppSTL) over $\bar{x}
  = \seq{x_1,\dots,x_n}$, and for all traces $\sigma \in (\R^n)^+$, it
  holds that:
  \begin{align*}
    \mon_\phi(\sigma) = \top \ (\text{resp., } \bot)
      \quad \Iff \quad 
    \sigma \models \phi \ (\text{resp., } \not\models\phi)
  \end{align*}
\end{restatable}
\begin{proof}
  We begin with the case of \FppSTL. Let $\ltl{F(\psi)}$ be a \FppSTL
  formula, over the variables $\bar{x} = \seq{x_1,\dots,x_n}$.
  First, notice that the set of good prefixes of $\ltl{F(\psi)}$ is exactly
  the set of \emph{models} of $\ltl{F(\psi)}$. This is proved in the claim
  below.

  \begin{claim}
    The set of good prefixes of $\ltl{F(\psi)}$ is exactly the language of
    $\ltl{F(\psi)}$.
  \end{claim}
  \begin{claimproof}
    To prove that every good prefix of $\ltl{F(\psi)}$ is a model of
    $\ltl{F(\psi)}$ it suffices to notice that,
    by~\cref{def:rt:mv:cosafety:lang}, a good prefix $\sigma$ is such that
    $\sigma\cdot\sigma'\models \ltl{F(\psi)}$, for all $\sigma' \in
    (\R^n)^*$. For the particular case in which $\sigma' = \epsilon$ (i.e.,
    the empty trace), it follows that $\sigma \models \psi$, and thus that
    $\sigma \models \ltl{F(\psi)}$.

    To prove the opposite direction, consider a model $\sigma$ of
    $\ltl{F(\psi)}$. By definition of the $\ltl{F}$ operator and since
    $\psi$ is a formula of \ppSTL, it holds that there exists a prefix
    $\sigma'$ of $\sigma$ such that $\sigma'\cdot \sigma'' \models \psi$,
    for all $\sigma'' \in (\R^n)^*$. Since $\sigma'$ is a prefix of
    $\sigma$, this implies that $\sigma \cdot \sigma'' \models
    \ltl{F(\psi)}$, for all $\sigma'' \in (\R^n)^*$. Therefore, $\sigma$ is
    a good prefix of $\ltl{F(\psi)}$.
  \end{claimproof}

  Now, we prove that $\mon_\phi(\sigma) = \top$ iff $\sigma \models \phi$.
  Let $\sigma \in (\R^n)^+$ be a trace.  By definition of $\mon_\phi$
  (~\cref{sub:monitoring}), we have that $\mon_\phi(\sigma) = \top$ iff
  $\sigma\cdot\sigma' \models \phi$, for all $\sigma' \in (\R^n)^*$. This is
  equivalent of saying that $\sigma$ is a \emph{good prefix} of $\phi$
  (\cref{def:rt:mv:cosafety:lang}). By the claim above, this is true iff
  $\sigma$ is a model of $\phi$, that is $\sigma \models \phi$. This
  concludes the proof for the case of \FppSTL.

  The case of \GppSTL is obtained analogously by exploiting the fact that
  the language of $\ltl{G(\psi)}$ is exactly the complement of the language
  of $\ltl{F(!\psi)}$ and the fact that the bad prefixes of $\ltl{G(\psi)}$
  are the good prefixes of $\ltl{F(!\psi)}$.
\end{proof}

It follows that the monitoring problem of a trace $\sigma$ and a formula
$\phi$ of \GppSTL or of \FppSTL can be solved in time $\O(|\sigma|
\cdot |\phi|)$.
In the next section, we exploit this result to develop a GPU-based
trace checking module that is used for efficient early failure detection.

\section{Failure detection framework}
\label{sec:framework}

Here, we describe the multiple elements that compose the framework we propose, i.e., the GPU-based module that relies on trace checking to perform \STL monitoring, the framework training algorithm, and the evolutionary algorithm (EA) for formula extraction. Again, we focus on the specific case of finite and discrete traces. 

\subsection{GPU-based \STL monitoring by trace checking}
\label{sub:stl_vect}

Consider $T \subseteq (\R^n)^+$, $|T| = m$, and let $l = \max_{\tau \in T} |\tau|$. We can represent $T$ as the matrix $\mathbf{T} \in \mathbb{R}^{m \times l \times n} $, padding traces shorter than $l$. 
Suppose we have a formula $\phi \in \ppSTL$. To perform trace checking, we consider the \textit{parse tree} of $\phi$, where every internal node is labeled with a temporal or boolean operator according to the subformula of $\phi$ it represents, and leaf nodes are \textit{atomic formulas}.

We adapted the algorithm proposed for \LTL over finite traces in \cite{fionda2016complexity}, considering every node as a function returning a \textit{robustness} value 
according to the quantitative semantics of \STL. 
Recall that the robustness $\rho(\phi,\tau,i)$ of formula $\phi$ for a trace $\tau$ at a time point $i$ tells us how much (resp. far) $\phi$ is (resp. is from being) satisfied on $\tau$ at $i$.
To leverage GPU acceleration, we implemented 
$\rho(\cdot)$
by means of vectorized functions, each returning a \textit{vector} $rob_{\tau, \phi} \in \R^l$ \textit{of robustness values}, one for every time point in the trace (i.e., $rob_{\tau, \phi}[j] = \rho(\phi,\tau,j)$ for $0\leq j < l$).
Intuitively, leaf nodes are of type $\mathbb{R}^{l \times n} \to \mathbb{R}^{l }$, while internal nodes are $\mathbb{R}^{l } \to \mathbb{R}^{l }$ or $\mathbb{R}^{l } \times \mathbb{R}^{l } \to \mathbb{R}^{l }$, according to their arity.

Considering $\mathbf{T}$ rather than $\tau$, each vectorized function can be straightforwardly applied to a matrix (representing a set of traces) rather than a single vector ($\mathbb{R}^{m \times l \times n} \to \mathbb{R}^{m \times l }$, $\mathbb{R}^{m \times l } \to \mathbb{R}^{m \times l }$, $\mathbb{R}^{m \times l } \times \mathbb{R}^{m \times l } \to \mathbb{R}^{m \times l }$). 
Thus, formula $\phi$ is interpreted as a %
function $g_\phi : \mathbb{R}^{m \times l \times n} \to \mathbb{R}^{m \times l }$ (that returns a robustness matrix $rob_{\mathbf{T}, \phi} \in \mathbb{R}^{m \times l }$, where $rob_{\mathbf{T}, \phi}[k,j] = \rho(\phi,\mathbf{T}[k],j)$ for $0\leq j < l$ and $0\leq k < m$), defined by inductively applying the vectorized functions from the leaves up to the root of $\phi$'s parse tree. %
We further increased the parallelization of our algorithm by simultaneously considering multiple formulas. %
Given $\phi_1, \dots, \phi_r \in \ppSTL$, we build the parse tree for $ \Phi \coloneqq \bigwedge_{1 \leq i \leq r} \phi_i$, and stop the bottom-up evaluation at the first level of the tree rather than the root, leading to $g_\Phi : \mathbb{R}^{m \times l \times n} \to \mathbb{R}^{r \times m \times l }$.

\begin{figure}[t]
    \centering
    \includegraphics[width=.83\linewidth,clip,trim={5.5em 2.em 2.2em 1.63em}]{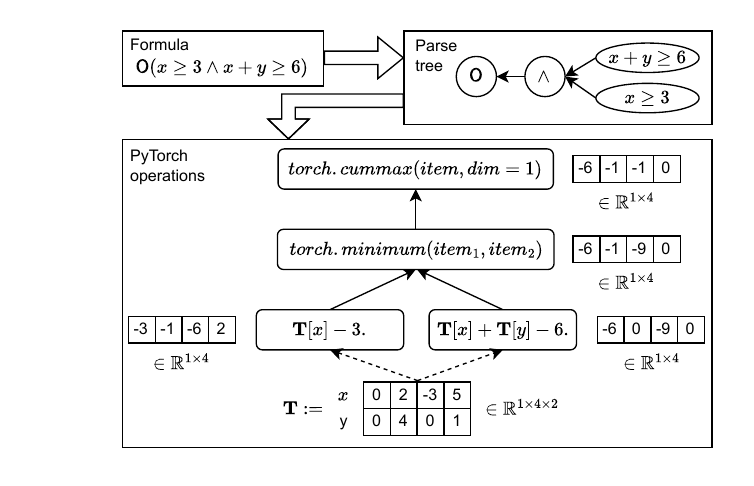}
    \caption{An intuitive example of our trace checking implementation for $\phi = \mathsf{O}(x \geq 3 \wedge x + y \geq 6)$ and a trace with 2 variables and 4 time instants.}
    \label{fig:parser_vec_exaple}
\end{figure}

An %
example of our trace checking implementation for the formula $\phi = \ltl{O(x \geq 3 \wedge x + y \geq 6)}$ and a single bi-variate trace contained in a matrix $\mathbf{T}$ is reported in \cref{fig:parser_vec_exaple}. %
It shows how the robustness vector/matrix, say $rob_{\mathbf{T}, \phi}${}, %
is computed. 
The last step is to consider $\ltl{G}(\neg \phi) \in \GppSTL$ (resp., $\ltl{F}(\phi) \in \FppSTL$). Then, the monitoring output is obtained as: $\mon_{\ltl{G}(\neg \phi)} = torch.cummin((-rob_{\mathbf{T}, \phi}) \geq 0, dim = 1) \in \{\bot, \top\}^{1 \times 4}$ (resp., $\mon_{\ltl{F}(\phi)} = torch.cummax((rob_{\mathbf{T}, \phi}) \geq 0, dim = 1) \in \{\bot, \top\}^{1 \times 4}$).

The detailed implementation of all vectorized functions modeling the quantitative semantics of \STL, along with our GPU-accelerated trace checking approach, can be found in the code provided in~\cref{app:reprod}.
Our current implementation serves as an experimental proof of concept and leverages Python partial functions, PyTorch~\cite{NEURIPS2019_9015}, and a parser based on Lark~\cite{lark_parser}. 
In the future, we plan to release a more efficient version, %
incorporating a dedicated component for online monitoring that utilizes dynamic programming.

\subsection{Learning temporal specifications}
\label{sub:training}

\begin{algorithm}[t]
\SetAlgoLined
\DontPrintSemicolon
\fontsize{8}{6.75}\selectfont
\SetKwInOut{Input}{input}
\caption{Framework training phase}\label{alg:algorithm_training}
\Input{%
$\mathcal{P}$ initial (possibly empty) pool of formulas, %
$\mathcal{X}$ dataset of labelled traces $(\sigma, is\_failure)$, %
\quality quality requirements, %
$e$ number of epochs, %
$b$ batch size}
$\mathcal{X} \gets \textsc{normalizeTraces}(\mathcal{X})$ \;
$\Sigma_\top \gets \{ \sigma \suchthat (\sigma, is\_failure) \in \mathcal{X} \land is\_failure = \top \}$\;
$\Sigma_\bot \gets \{ \sigma \suchthat (\sigma, is\_failure) \in \mathcal{X} \land is\_failure = \bot \}$\;
$\Sigma_\bot \gets \Sigma_\bot \cup \textsc{genAugmentedTraces}(\Sigma_\bot)$\;
$F \gets \emptyset$\;
\For{$\sigma \in \Sigma_\top$}{
    $A_\sigma \gets  \textsc{genAugmentedTraces}(\{\sigma\})$\;
    $F \gets F \cup \{(\sigma, A_\sigma)\}$ \;
}

\For{$e$ times}{
        $F \gets \{\textsc{cutTraces}((\sigma, A_\sigma), \mathcal{P}) \suchthat (\sigma, A_\sigma) \in F\}$\;
    $B \gets \textsc{generateBatches}(F, b)$\;
    \For{$\beta \in B$}{
        $\Phi \gets \textsc{learnFormulas}(\beta, \Sigma_\bot, q)$ \;
        $\mathcal P \gets \mathcal P \cup \{\ltl{G}(\neg \phi) \suchthat \phi \in \Phi \}$ \;
    }
}

\Return{$\mathcal{P}$} \;

\end{algorithm}

\Cref{alg:algorithm_training} summarizes the framework training procedure, which, given a training dataset, learns a set of formulas for early failure detection on trace data.
Its input consists of:
\begin{inparaenum}[(i)]
    \item a pool $\mathcal{P}$ of formulas encoding behaviors that must always be followed by the system, which may initially be empty or 
    prefilled by domain experts;
    \item a training set $\mathcal{X}$ composed of pairs $(\sigma, is\_failure)$, where $\sigma$ is a system execution trace and $is\_failure$ is a Boolean value indicating whether the trace ends in failure ($\top$) or not ($\bot$);
    \item a set of quality requirements \quality that any newly learned formula must satisfy; 
    \item the number of training epochs \epoch;
    \item the batch size \batch.
\end{inparaenum}
The output of \Cref{alg:algorithm_training} is the %
pool $\mathcal{P}$ enriched with monitoring formulas learned from set $\mathcal{X}$.

The training starts by normalizing input traces into the interval $[0,1] \subset \mathbb{R}$ (\Cref{alg:algorithm_training}, Line~1), and extracting the subset of failure (resp., good) traces $\Sigma_\top$ (resp., $\Sigma_\bot$) $\mathcal{X}$ (Lines~2--3). 
Then, for each good and failure trace, 
augmented traces are created by injecting random Gaussian noise (to counter overfitting). The set $\Sigma_\bot$ is extended with $\Sigma_\bot$'s augmentations; and, the set $F$ is generated, which stores each failure trace $\sigma$ paired with its augmentations $A_\sigma$ (Lines~4--8). 

The iterative part of the algorithm consist of \epoch passes (epochs) over the training set (Lines~9--14). As a first step (Line~10), all failure training traces $(\sigma,A_\sigma) \in F$ are trimmed on the right. 
To do this, we perform trace checking based on the monitoring pool $\mathcal{P}$, which consists, by construction, of safety formulas of the kind $\psi \in \GppSTL$; thus, $\psi$ leads to a violation/failure if $\mon_\psi(\sigma) = \bot$. 
Given
$v_\sigma \coloneqq \min_{\psi \in \mathcal{P}}\min\{i \suchthat \rho(\psi, \sigma, i) < 0, 0 \leq i < |\sigma|\}$
the earliest violation point in $\sigma$ for all formulas in $\mathcal{P}$, if it exists, we restrict $\sigma$ to the prefix $\sigma_{[0,v_\sigma-1]}$, and the same $v_\sigma$ is used to restrict all the augmentations $A_\sigma$. This part is crucial, as it allows for the progressive extraction of formulas that increasingly anticipate the failures with each epoch.

Each epoch processes multiple batches of traces, with formulas extracted from each batch.
First, $\lceil |F| \div \batch \rceil$ non-overlapping batches are created for the current epoch, where each batch is a random subset of $F$ containing \batch pairs of the kind $(\sigma, A_\sigma)$ (Line~11). 
For every batch $\beta \in B$ (Line~12), the function \textsc{learnFormulas} attempts to extract, for each non-augmented failure trace $\sigma \in \beta$, a formula $\phi$ capable of identifying a bad behavior of the system.
A possibly empty formula set $\Phi$, $|\Phi| \leq |\beta|$, is produced  (Line~13), which contains all formulas $\phi$ that satisfy predefined quality criteria \quality. These ensure effective detection of failure behavior in $\beta$ while keeping the False Alarm Rate (the fraction of false failure detections) with respect to $\Sigma_\bot$ below a specified threshold.
This extraction process is carried out by our newly proposed EA, which relies on trace checking, %
enabling GPU-based parallelization. %
Finally, the safety $\GppSTL$ rewriting of each learned formula is added to the pool (Line~14).\footnote{This stems from the fact that, due to the encoding of its fitness function, the EA aims at generating formulas $\phi\in\Phi$ that identify (evaluating to $\top$) bad behaviours on trace suffixes.}

For the sake of simplicity, in our algorithm the batches $\beta \in B$ are processed sequentially (Lines~12--14). However, since the batches are independent from each other, the procedure can be scaled to process multiple batches in parallel if multiple GPUs or, more generally, multiple accelerators (e.g., TPUs) are available.

During \Cref{alg:algorithm_training} the pool of formulas $\mathcal{P}$ is iteratively 
expanded and returned as the output (Line~15). 

Note that, thanks to the inherent interpretability of the formulas, domain experts can, in principle, 
inspect and intervene at any time to modify the pool $\mathcal{P}$; for example, by manually specifying a new formula or removing/modifying one that is unsubstantiated according to their knowledge.

\subsection{EA-based formula extraction}
\label{sub:genetic}

The function \textsc{learnFormulas} is implemented using a multi-objective EA which performs a genetic programming task, aiming to evolve formulas from an initial population of random solutions~\cite{poli2008gp}. To speed up the fitness evaluation step, the algorithm employs  monitoring via trace checking, leveraging GPU acceleration, as described previously. The implementation is carried out using the DEAP (Distributed Evolutionary Algorithms in Python) library~\cite{DEAP_JMLR2012}. 

\begin{figure}[t]
    \centering
\includegraphics[width=0.9\linewidth,clip,trim={0 0 0 0}]{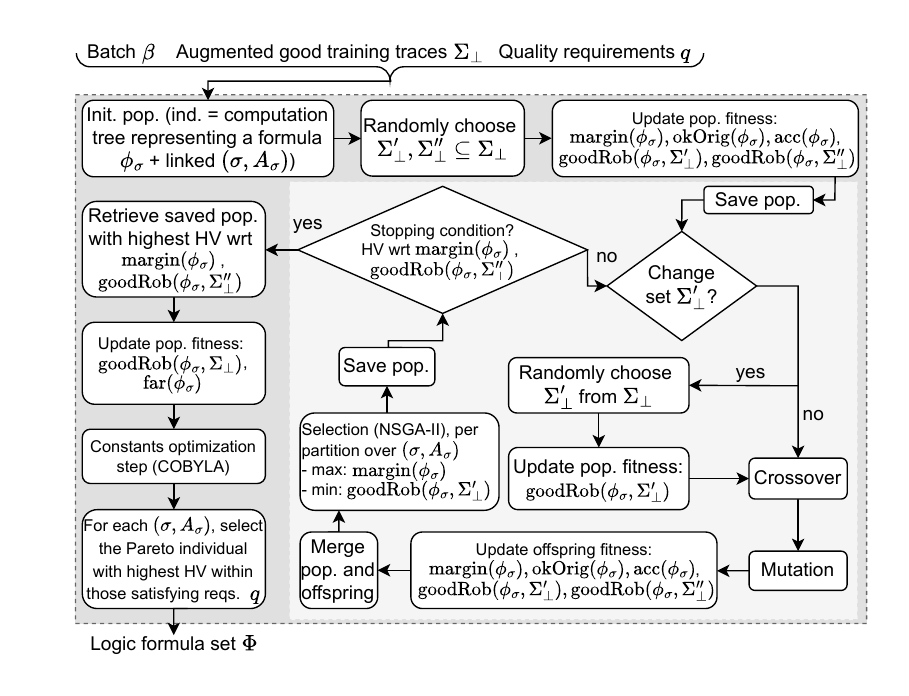}
    \caption{General workflow of the EA for formula extraction.}
    \label{fig:genetic}
\end{figure}

\Cref{fig:genetic} outlines the EA workflow.  The algorithm takes as input a batch $\beta$ of pairs consisting of failure traces and their augmentations $(\sigma, A_\sigma)$, the set of augmented good training traces $\Sigma_\bot$, and the set of quality requirements \quality{} that the output formulas must satisfy.  
As for its output, it produces a set $\Phi$ that contains at most one formula $\phi_\sigma$ for each trace $\sigma$. The idea is that 
formula $\phi_\sigma$ captures an anticipatory bad behavior in $\sigma$, evaluating to $\top$ on one of its suffixes.

\paragraph*{Population and its initialization}

First we initialize the population. 
Each individual encodes a random \ppSTL formula $\phi_\sigma$ represented as a computation tree. %
The leaves of the computation tree may represent either a variable corresponding to one of the signals in a multivariate trace or a constant $c \in [0,1] \subset \mathbb{R}$, while the internal nodes encode the logical operators of \ppSTL. Tree heights range from 1 to 17 \cite{koza1994genetic}. The upper limit of 17 is always maintained through the EA execution. 
The formula $\phi_\sigma$ is randomly linked with one of the pairs $(\sigma, A_\sigma)$ in the input batch, which will be considered for its fitness evaluation. 
As we shall see, an individual thus encodes a candidate formula, potentially capable of performing early failure detection over the traces in $\{\sigma\} \cup A_\sigma$.

\paragraph*{Fitness of the individuals}

The fitness of $\phi_\sigma$ is composed by six %
quantities, calculated and tracked at different steps of the EA. 

The first three quantities are associated with the corresponding pair $(\sigma, A_\sigma)$, aiming to evaluate how well $\phi_\sigma$ separates traces in $\{\sigma\} \cup A_\sigma$ into good behaviour prefixes and failure suffixes, finding the best possible splitting point. 
Given a trace $t \in \{\sigma\} \cup A_\sigma$ and $\phi_\sigma$, we perform trace checking (for simplicity, we describe the EA operation for a single $\sigma$ and $\phi_\sigma$, but in practice we compute the robustness for all traces and formulas in the population with a single, simultaneous vectorized operation on the GPU) obtaining an array of robustness values $rob_{t, \phi_\sigma}$ s.t. $|t| = |rob_{t, \phi_\sigma}| = l$.\footnote{Since we normalize traces, $\text{robustness} \in [-1,1] \subset \mathbb{R}$ in our case.}
Given \(\text{maxRob}(t, \phi_\sigma) := \max_{0 \leq k < |t|} (rob_{t,\phi_\sigma} [k])\), we define:%
\begin{align*}
    score_{t, \phi_\sigma}[i] \coloneqq \min (-&\tanh (\text{maxRob}(t_{[0,i-1]}, \phi_\sigma)), \\ & \tanh(rob_{t, \phi_\sigma}[i])) 
\end{align*}
for all $0 \leq i \leq l$. Intuitively, $score_{t, \phi_\sigma}$ stores the \lq\lq goodness\rq\rq{} values (higher is better) assigned to each possible splitting point $i$ in the trace $t$.\footnote{For the two cases when no prefix or suffix are produced by the split point ($\epsilon \approx 0^+$): 
$score_{t, \phi_\sigma}[0]  \coloneqq \min(-\tanh(rob_{t, \phi_\sigma}[0] + \epsilon), \tanh(rob_{t, \phi_\sigma}[0]))$; 
$score_{t, \phi_\sigma}[l]  \coloneqq \min(-\tanh(\text{maxRob}(t_{[0,l-1]}, \phi_\sigma)), \tanh(\text{maxRob}(t_{[1,l-1]}, \phi_\sigma)))$.}
A good splitting point $i$ is one where $rob_{t, \phi_\sigma}[i] \gg 0$, leading to a \lq\lq strong\rq\rq{} failure detection, while all preceding robustness values $rob_{t, \phi_\sigma}[0:i-1] \ll 0$, ensuring a clear separation between the identified good behavior prefix $t_{[0,i-1]}$ and the failure suffix $t_{[i,l-1]}$. The hyperbolic tangent function emphasizes robustness values near the 0 threshold. 
We can now define the first three fitness values:
\begin{align*}
    &\text{margin}(\phi_\sigma) \coloneqq \min\limits_{t\in \{\sigma\} \cup A_\sigma}(\max\limits_{0 \leq k \leq |t|} (score_{t, \phi_\sigma}[k])) \\
    &\text{okOrig}(\phi_\sigma) \coloneqq rob_{\sigma,\phi_\sigma} [0] < 0 \land \text{maxRob}(\sigma, \phi_\sigma) \geq 0 \\
    &\text{acc}(\phi_\sigma) \coloneqq \frac{\sum_{t\in \{\sigma\} \cup A_\sigma} \mathds{1}(rob_{t,\phi_\sigma} [0] < 0 \land \text{maxRob}(t, \phi_\sigma))}{|\{\sigma\} \cup A_\sigma|} 
\end{align*}
where $\mathds{1}$ is the indicator function, giving a value of 1 to $\top$ and 0 to $\bot$. 
Intuitively, $\text{margin}(\phi_\sigma) \in [-1,1] \subset \mathbb{R}$ should be as high as possible and tells how good, in terms of separation measured in robustness, $\phi_\sigma$ is in splitting the failure traces $(\sigma, A_\sigma)$ into a good behaviour prefix and a failure suffix, considering the worst case; $\text{okOrig} (\phi_\sigma) \in \{\top, \bot\}$, should equal to $\top$, and tells if $\phi_\sigma$ 
splits the original failure trace $\sigma$ into a good prefix and a bad suffix;
and $\text{acc}(\phi_\sigma) \in [0,1]$ should be as high as possible and equals to the fraction of failure traces $(\sigma, A_\sigma)$ %
splitted by $\phi_\sigma$ into a good prefix and a bad suffix.

The other three fitness values refer to the sets $\Sigma_\bot$, $\Sigma_\bot'$ and $\Sigma_\bot''$. The latter two sets are randomly sampled from $\Sigma_\bot$ such that their sizes satisfy \( |\Sigma'_\bot| = |\Sigma''_\bot| = \fractgood \cdot |\Sigma_\bot| \), with \( \fractgood \in [0,1] \) framework global hyperparameter. We define:
\begin{align*}
    &\text{far}(\phi_\sigma) \coloneqq \frac{\sum_{t \in \Sigma_\bot} \mathds{1}(\text{maxRob}(t, \phi_\sigma) \geq 0)}{|\Sigma_\bot|} \\
    &\text{goodRob}(\phi_\sigma, S) \coloneqq \max_{t\in S}(\text{maxRob}(t, \phi_\sigma)) 
\end{align*}
where $S \in \{\Sigma_\bot', \Sigma_\bot''\}$. The function $\text{far}(\phi_\sigma)$ computes the false alarm rate over the traces in $\Sigma_\bot$, which should be as low as possible, since it represents the fraction of good traces mistaken as failure ones; the function $\text{goodRob}(\phi_\sigma, S)$ tells us, in terms of robustness, how distant $\phi_\sigma$ is from mistaking a good trace (of $\Sigma_\bot'$ or $\Sigma_\bot''$) as a failure one, in the worst case (thus, lower is better). 

As for how the six quantities are used in the EA,  $\text{margin}(\phi_\sigma)$ and $\text{goodRob}(\phi_\sigma, \Sigma_\bot')$ guide the evolutionary process as part of the EA's bi-objective fitness function, and are respectively to be maximized and minimized. 
Then, again $\text{margin}(\phi_\sigma)$ together with $\text{goodRob}(\phi_\sigma, \Sigma_\bot'')$ are used to implement an early stopping strategy. 
Finally, $\text{acc}(\phi_\sigma)$, $\text{okOrig}(\phi_\sigma)$ and $\text{far}(\phi_\sigma)$ are considered alongside the quality requirements $q$, which a formula must meet to be returned by the EA. %
Early stopping and quality criteria are described
later. %

\paragraph*{Crossover and mutation}

Given two randomly selected individuals $\phi_\sigma$ and $\phi'_{\sigma'}$, 
the crossover operation can hybridize them in two ways, with equal probability. 
In the first case, following the method described in \cite{brunello2024learning}, a crossover point is randomly chosen within each formula’s computation tree, and the subtrees rooted at such points are exchanged. In the second case, the reference traces associated with the two formulas are swapped, resulting in two offspring $\phi_{\sigma'}$ and $\phi'_\sigma$.

As for the mutation, given a randomly selected individual, three operations can be performed with equal probability, similarly to 
\cite{brunello2024learning}. All of them involve modifications to the computation tree of the formula: an internal node may be replaced by a randomly chosen operation; %
the overall tree may be shrunk by choosing randomly a branch and replacing it with one of its arguments, also randomly chosen; %
or, the values of %
constants in the tree may be changed.%

\paragraph*{Selection}

At each generation $g$, the population consists of the individuals inherited from generation $g-1$, along with the offspring created from them at generation $g$ through the crossover and mutation operators. To determine which individuals will advance to the next generation $g+1$, we first partition the individuals based on their reference set $(\sigma, A_\sigma)$ and process each partition separately. 
For each partition, selection is performed using NSGA-II \cite{Deb:2002:FEM:2221359.2221582}, ensuring that the size of the partition at generation $g+1$ matches its size at generation $g-1$. Recall that the objectives for selection are to maximize $\text{margin}(\phi_\sigma)$ and to minimize $\text{goodRob}(\phi_\sigma, \Sigma_\bot')$.

\paragraph*{Early stopping and post-evolutionary optimization}

The evolution %
terminates under one of two conditions: either the maximum number of generations $\maxgen$ (a framework's global hyperparameter) is reached, or the early stopping condition is triggered. 
The latter is defined as follows: the population hypervolume \cite{guerreiro2021hypervolume}, calculated with respect to $\text{margin}(\phi_\sigma)$ and $\text{goodRob}(\phi_\sigma, \Sigma_\bot'')$, is tracked across generations. If no improvement in the hypervolume is observed for \patience (an hyperparameter) generations, the execution is halted.

Once the evolutionary part has terminated, the population with the highest hypervolume, calculated as in the early stopping method, is retrieved. For each individual $\phi_\sigma$ in this population, we compute $\text{goodRob}(\phi_\sigma, \Sigma_\bot)$ and $\text{far}(\phi_\sigma)$. Next, the population is partitioned according to $(\sigma, A_\sigma)$. 
Within each partition, we select the top-$\kopt$ individuals (with $\kopt = 5$ by default) based on the hypervolume metric calculated using $\text{margin}(\phi_\sigma)$ and $\text{goodRob}(\phi_\sigma, \Sigma_\bot)$. These individuals undergo an additional optimization step to refine the constants in their computation tree, using the Constrained Optimization BY Linear Approximation (COBYLA) algorithm as implemented in the Scipy library \cite{Powell1994,2020SciPy-NMeth}, capping the iterations to 50. Finally, for each partition, we add to the return set $\Phi$ of \textsc{learnFormulas} the individual $\phi_\sigma^{\text{best}}$ (if any) with the highest hypervolume among those satisfying the quality criteria $q$. The latter are defined as $\text{acc}(\phi_\sigma^{\text{best}}) \geq \minacc$, $\text{okOrig}(\phi_\sigma^{\text{best}}) = \top$, and $\text{far}(\phi_\sigma^{\text{best}}) \leq \maxfar$ (see \cref{app:hp_search} for all the hyperparameters of the framework).

\subsection{Comparison to related work}

Most contributions in the literature extract temporal relations from time series data~\cite{aggarwal2018two,gao2020task,lu2020making,petmezas2021automated,wen2024new}, often relying on black-box techniques, neglecting the crucial aspects of interpretability and formality. Given that runtime verification has a long tradition of application in critical safety scenarios~\cite{leucker2009brief}, we believe interpretability by design, grounded in a well-defined formal framework, is paramount, even more in light of the %
limitations of post-hoc explanations~\cite{bilodeau2024impossibility}.

Some papers follow a different approach, focusing on learning \LTL specifications that separate two set of traces defined over a fixed alphabet \cite{DBLP:conf/birthday/Neider025}.
The problem is \NP-HARD \cite{DBLP:conf/icgi/FijalkowL21}, and state-of-the-art methods struggle to scale \cite{DBLP:conf/cav/ValizadehFB24}. To solve this problem, some authors  propose a %
GPU-based enumeration algorithm for \LTL learning
\cite{DBLP:conf/cav/ValizadehFB24}. Yet, it uses bit vector representations and bit wise operations, and is not directly applicable to \STL and multivariate real-valued traces.

To the best of our knowledge, the only contributions that propose to combine monitoring and machine learning for early 
failure detection %
are \cite{BrunelloMMSU23,brunello2024learning}. 
Our work substantially differs from them.
Theoretically, we focus on making monitoring computationally more tractable, while preserving the expressive power. %
Practically, we propose a %
quite different algorithm, leading to much better performance. %
Key modifications are: 
\begin{inparaenum}[(i)]
\item %
a new bi-objective fitness function for the EA, calculated by means of trace checking and GPU acceleration, which significantly expedite the formula learning process;
\item thanks to the much faster candidate formula evaluation procedure, the design of a framework that incorporates the concepts of epochs and batches, which addresses one of the key limitations of~\cite{brunello2024learning} where the %
approach may learn from each training trace only once. This %
resulted in a waste of %
training data and a strong dependence on the order in which the algorithm observed the traces; %
\item %
a constants optimization step after the evolutionary phase of formula extraction;
\item %
formulas comparing signals, not just a signal with a constant.
\end{inparaenum}

\section{Experimental evaluation}
\label{sec:exper}

We applied our framework to three %
datasets from the literature, comparing its performance with %
recent contributions. 
We utilized the Backblaze Hard Drive,\footnote{\pdfulink{https://www.backblaze.com/b2/hard-drive-test-data.html}} Tennessee Eastman Process (TEP),\footnote{\pdfulink{https://doi.org/10.7910/DVN/6C3JR1}} and NASA C-MAPSS\footnote{\pdfulink{https://data.nasa.gov/dataset/C-MAPSS-Aircraft-Engine-Simulator-Data/xaut-bemq}} datasets. %

The \emph{Backblaze Hard Drive} dataset provides ongoing monitoring data of hard drives in the Backblaze data center, tracked using Self-Monitoring Analysis and Reporting Technology (SMART). Each trace includes the date of the report, the serial number of the drive, a failure label, and 21 numerical SMART parameters. For consistency with the literature, we focus on the ST4000DM000 model and adopt~\cite{brunello2024learning} Split~$S1$, which consists of a training set (October-November 2016, 11423 good drives and 148 failing drives) and a test set (December 2016, 23313 good drives and 86 failing drives). Due to time constraints, we exclude Split~$S2$, a larger but less challenging version of Split~$S1$, where failure detection performance in the literature achieves an F1 of ~0.93, compared to 0.56 for Split~$S1$.

The \emph{Tennessee Eastman Process} (TEP) dataset consists of simulated data from a fictitious chemical plant. It includes 1000 training and 1000 test traces, sampled every 3 minutes and labeled as Type~0 (normal behavior) or Type~1 (faulty behavior). Each training trace spans 25 hours, while each test trace lasts 48 hours, with 500 faulty traces in each set. Features include a trace ID, fault type, and 52 variables representing operational data from plant components.

The \emph{NASA Commercial Modular Aero-Propulsion System Simulation} (C-MAPSS) dataset provides run-to-failure simulated data of turbofan jet engines. In our analysis, we focus on the FD001 subset, which simulates engines operating under a single condition (\emph{Sea level}) with failures due to HPC degradation. Each simulation is a multivariate time series, sampled at one-second intervals from 21 engine sensors. The dataset contains 100 training traces (each ending in failure) and 100 test traces (each ending a specified number of steps before failure, referred to as the “gap”). Following \cite{kim2021multitask}, we target the detection of the \emph{Unhealthy state} class by labeling the 70\% prefix of each training trace as normal behavior and the full trace as faulty, generating 200 training traces. For the test set, traces are labeled as good (61 traces, if trace length $<$ 0.7*(length + gap)) or faulty (39 traces, if trace length $\geq$ 0.7*(length + gap)).

For each dataset, the framework is tuned using a separate validation set derived from the training data (details in \cref{app:hp_search}). The best hyperparameters obtained from this process are then applied to run the framework on the entire training set. To evaluate the framework, in an online fashion, on the test split of the dataset, the monitoring tool \texttt{rtamt}~\cite{nivckovic2020rtamt} is executed on each trace, using the pool of formulas $\mathcal{P}$ learned through \Cref{alg:algorithm_training}. %
To account for the stochastic nature of our approach, each experiment is repeated ten times. \cref{app:reprod} provides the code for reproducing our results and applying the framework to new scenarios.

\subsection{Results}

\begin{table}[t]
\centering
\caption{Experimental results.\label{tab:table_res}}
\resizebox{\linewidth}{!}{%
\begin{threeparttable}
\begin{tabularx}{0.675\textwidth}{p{3.3em}p{8.35em}p{0.7em}p{0.7em}p{1.2em}p{1.2em}p{1.2em}p{4.1em}p{1.2em}p{1.2em}p{3.35em}}
\toprule 
\textbf{Dataset} & \textbf{Approach} & \textbf{P} & \textbf{R} & \textbf{FAR} & \multicolumn{3}{c}{\textbf{F1}} & \multicolumn{3}{c}{\textbf{MCC}} \\
 & & & & & \textbf{Min} & \textbf{Max} & \textbf{Avg \stdf{std}}\cellcolor{gray!10} & \textbf{Min} & \textbf{Max} & \textbf{Avg\stdf{std}}\cellcolor{gray!10}\\
 \toprule
\multirow{4}{*}{\begin{tabular}[c]{@{}l@{}}Backblaze \\$S1$\end{tabular}}
 & \cite{huang2017hard} & $.51$ & $.54$ & $.00$ & - & - & $.52$\cellcolor{gray!10}  & - & - & - \cellcolor{gray!10} \\
 & \cite{lu2020making} & $.87$ & $.41$ & $.00$ & - & - & $.55$\cellcolor{gray!10}  &- & - & -\cellcolor{gray!10}  \\
 & %
 GP-\GppSTL ['24] & $.85$ & $.47$ & $.00$ & $.55$ & $.62$ & $.60$\stdf{.?}\cellcolor{gray!10}  & $.59$ & $.64$ & $.62$\stdf{.?}\cellcolor{gray!10}  \\
 & \textbf{Our solution} & $\mathbf{.87}$ & $.47$ & $.00$ & $\mathbf{.58}$ & $\mathbf{.64}$ & $\mathbf{.61}$\stdf{.02}\cellcolor{gray!10} & $\mathbf{.60}$ & $\mathbf{.66}$ & $\mathbf{.64}$\stdf{.02}\cellcolor{gray!10} \\
\midrule
\multirow{4}{*}{TEP}
 & \cite{hajihosseini2018fault} & $1.0$ & $1.0$ & - &  - & - & $1.0$\cellcolor{gray!10} &- & - & -\cellcolor{gray!10} \\
 & \cite{onel2019nonlinear} & $1.0$ & $1.0$ & $.00$ &  - & - & $1.0$\cellcolor{gray!10} &- & - & -\cellcolor{gray!10} \\
 & %
 GP-\GppSTL ['24] & $1.0$ & $1.0$ & $.00$ &  $1.0$ & $1.0$ & $1.0$\stdf{.00}\cellcolor{gray!10} & $.99$ & $1.0$ & $1.0$\stdf{.00}\cellcolor{gray!10} \\
 & \textbf{Our solution} & $1.0$ & $1.0$ & $.00$ &  $1.0$ & $1.0$ & $1.0$\stdf{.00}\cellcolor{gray!10} & $1.0$ &  $1.0$ &  $1.0$\stdf{.00}\cellcolor{gray!10} \\
\midrule
\multirow{3}{*}{C-MAPSS}
 & \cite{kim2021multitask} & $.71$ & $1.0$ & - &  - & - & $.83\cellcolor{gray!10}$ &- & - & -\cellcolor{gray!10}  \\
 & %
 GP-\GppSTL ['24] & $\mathbf{.98}$ & $.64$ & $\mathbf{.01}$ & $.71$ & $.87$ & $.77$\stdf{.05}\cellcolor{gray!10}  & $.63$  & $.82$ & $.70$\stdf{.06}\cellcolor{gray!10}  \\
 & \textbf{Our solution} & $.92$ & $\mathbf{.86}$ & $.05$ & $\mathbf{.85}$ & $\mathbf{.92}$ & $\mathbf{.89}$\stdf{.03}\cellcolor{gray!10} & $\mathbf{.77}$ & $\mathbf{.87}$ & $\mathbf{.83}$\stdf{.04}\cellcolor{gray!10} \\
\bottomrule
\end{tabularx}
\begin{tablenotes}
\small
\item \cite{lu2020making} results listed as in \cite{brunello2024learning}; GP-\GppSTL [2024] recomputed with the original code; others reported as in the original refs. P = precision; R = recall. 
\end{tablenotes}
\end{threeparttable}
}
\end{table}

\begin{figure*}[t]
  \centering
  \subfigure[Backblaze \(S1\) (%
  in days).\label{fig:anticip_smart}]{
    \includegraphics[width=0.32\linewidth,clip,trim={0 1em 0 1em}]{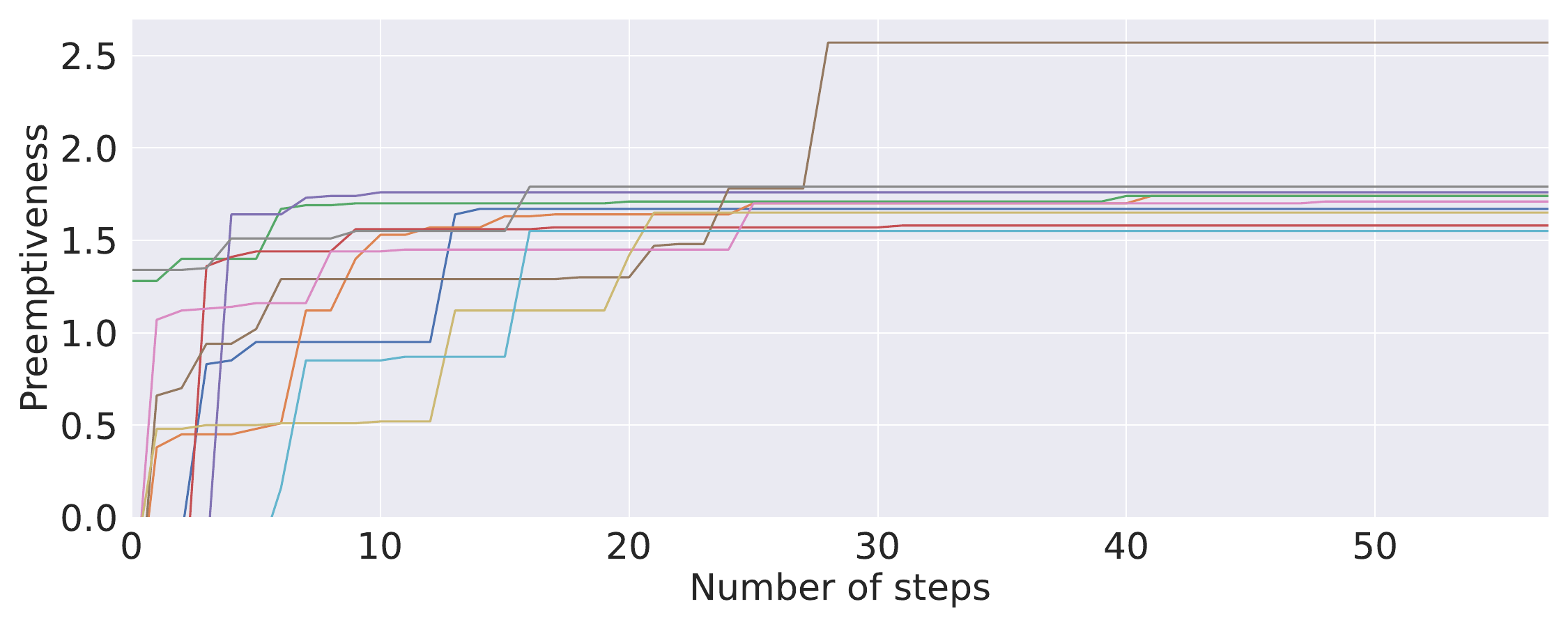}}
  \hfill
  \subfigure[TEP (pre-emptiveness in 3-min steps).\label{fig:anticip_tep}]{
    \includegraphics[width=0.32\linewidth,clip,trim={0 1em 0 1em}]{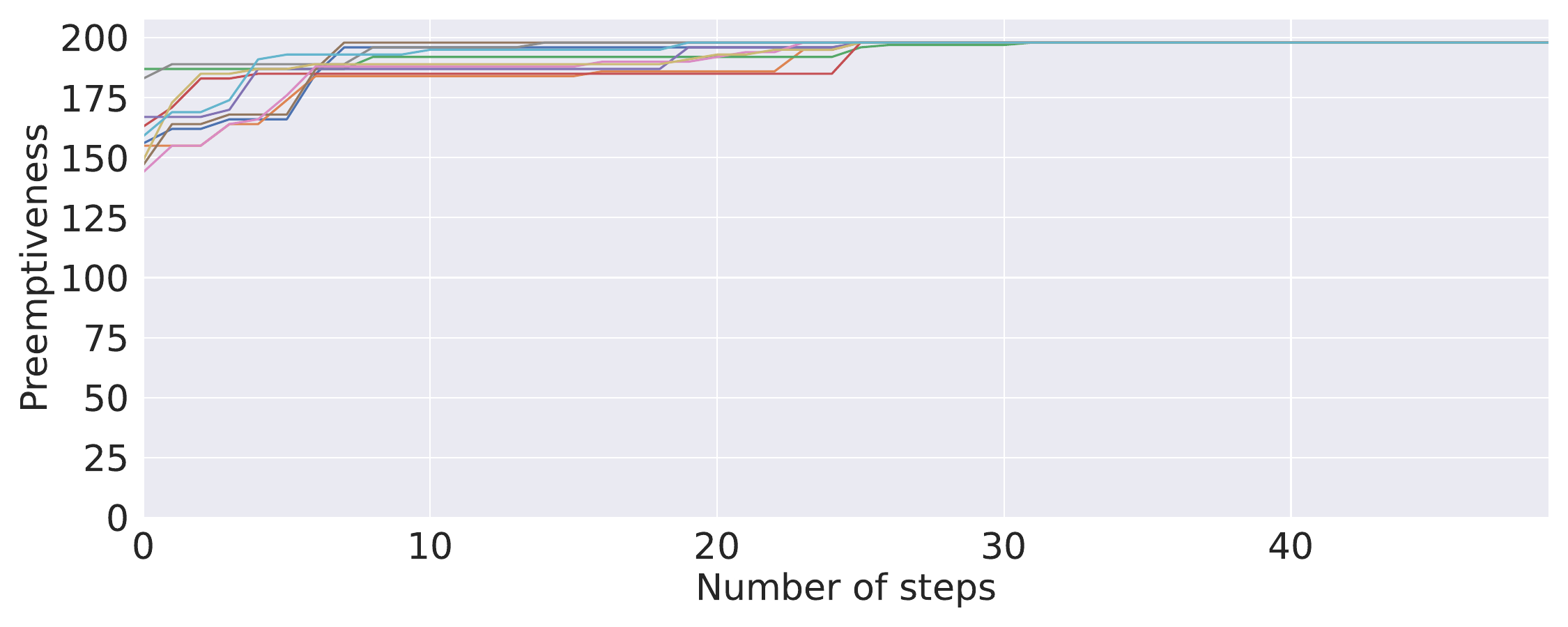}}
  \hfill
  \subfigure[C-MAPSS (%
  in seconds).\label{fig:anticip_cmapss}]{
    \includegraphics[width=0.32\linewidth,clip,trim={0 1em 0 1em}]{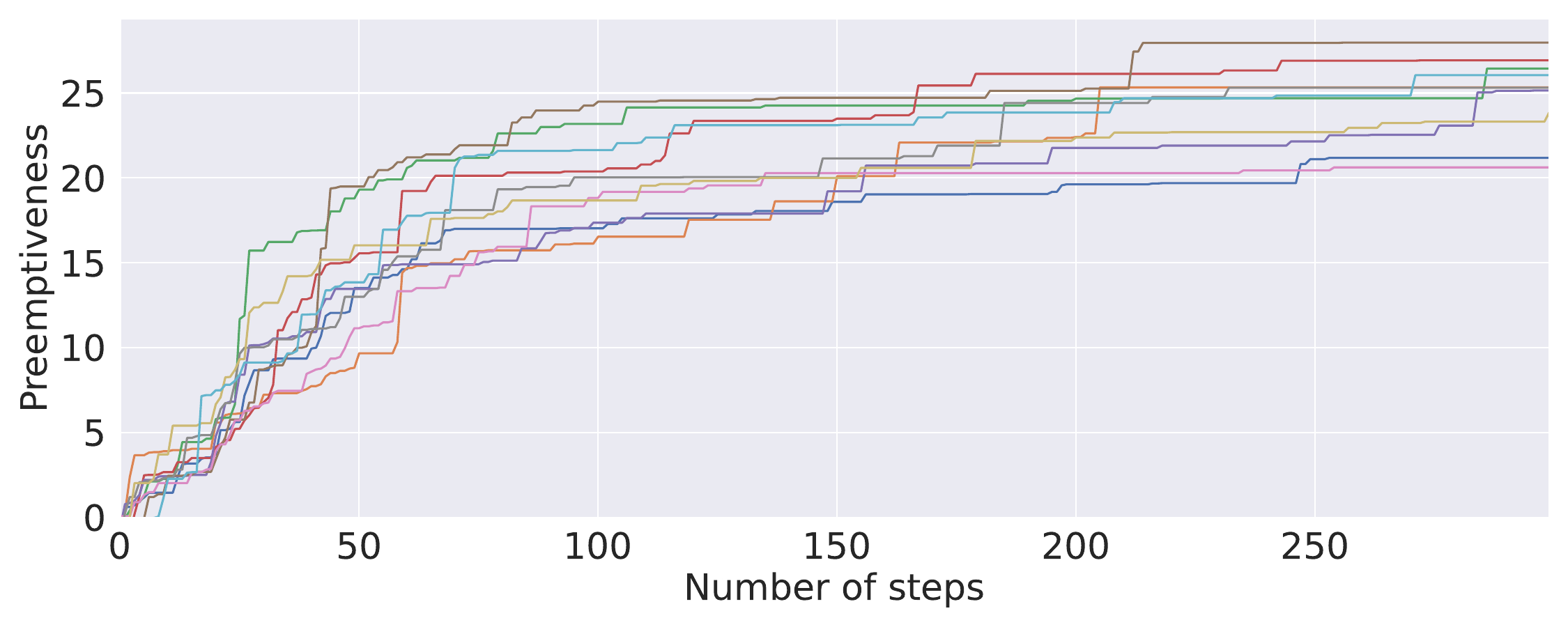}}
  \vspace{-.5em}
  \caption{Test set failure detection average preemptiveness, as additional properties are learned and incorporated into the pool (10 framework runs).}
  \label{fig:anticip}
\end{figure*}
\begin{figure*}[!ht]
  \centering
  \subfigure[Backblaze \(S1\).\label{fig:f1_smart}]{
    \includegraphics[width=0.32\linewidth,clip,trim={0 1em 0 1em}]{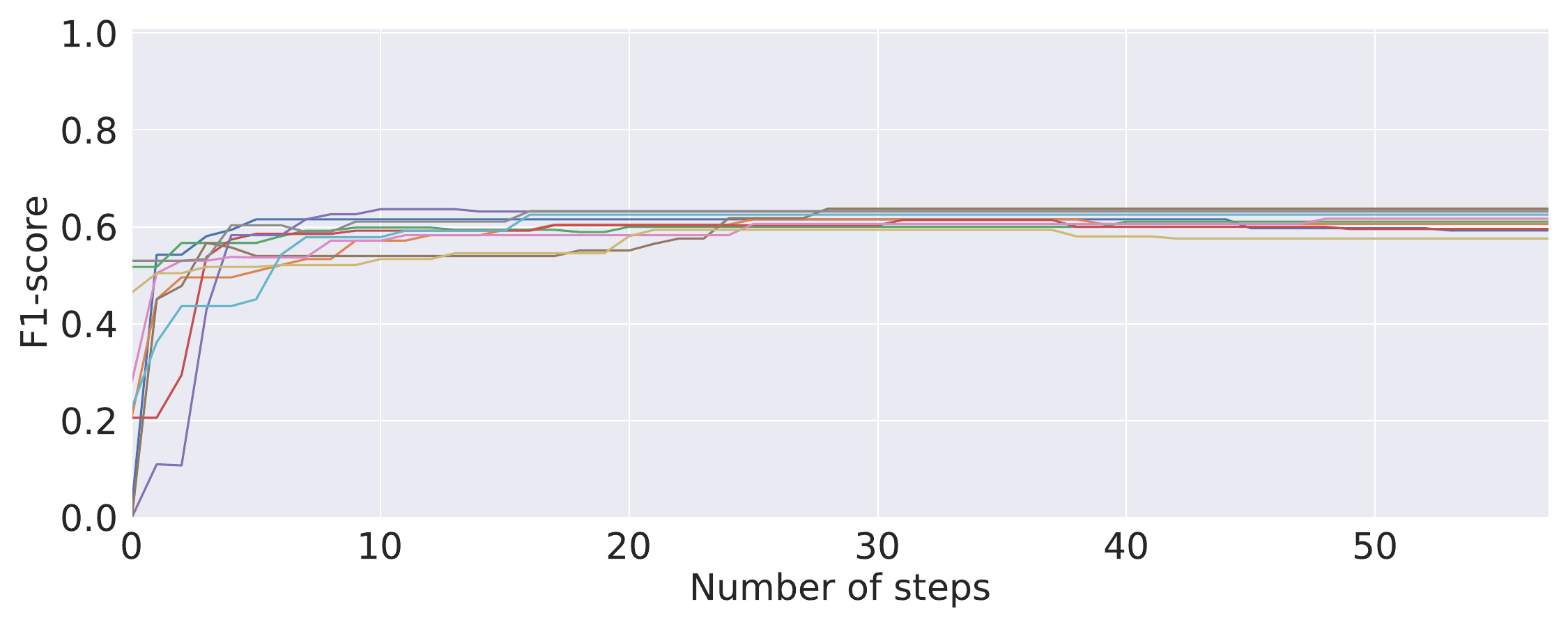}}
  \hfill
  \subfigure[TEP.\label{fig:f1_tep}]{
    \includegraphics[width=0.32\textwidth,clip,trim={0 1em 0 1em}]{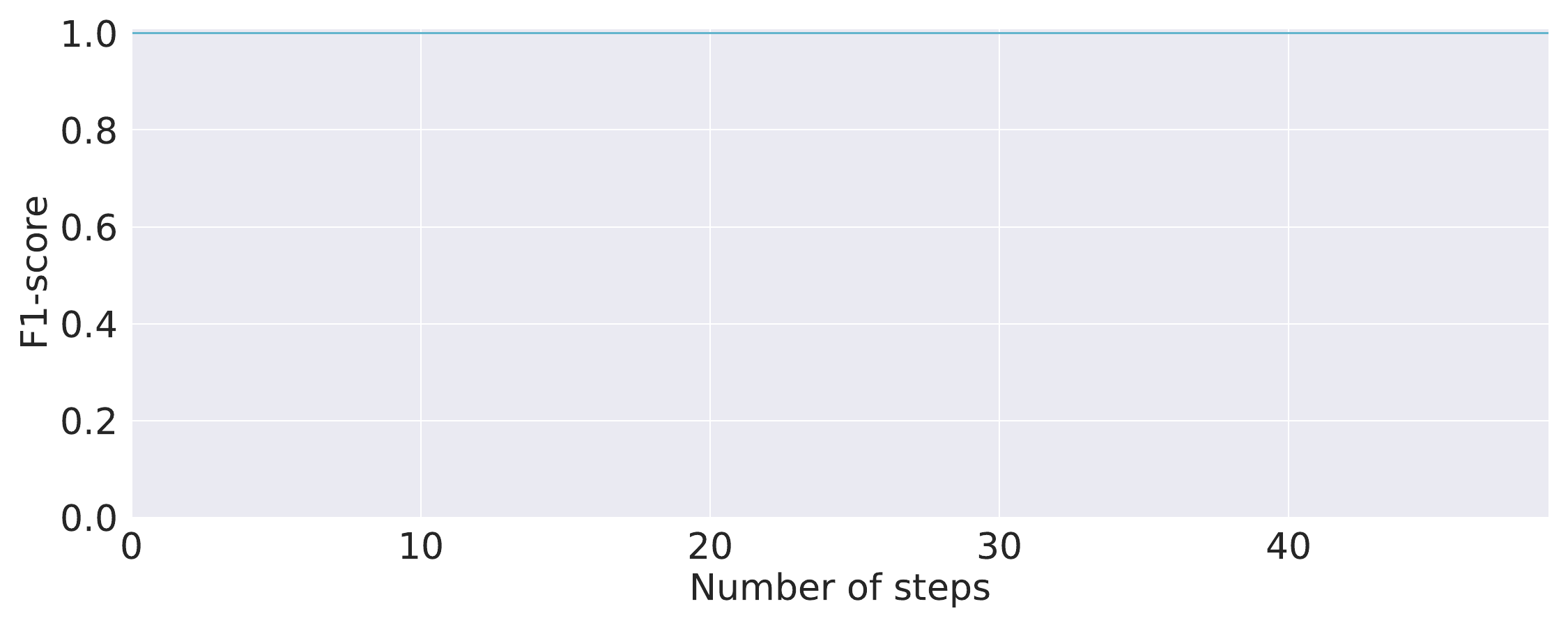}}
  \hfill
  \subfigure[C-MAPSS.\label{fig:f1_cmapss}]{
    \includegraphics[width=0.32\linewidth,clip,trim={0 1em 0 1em}]{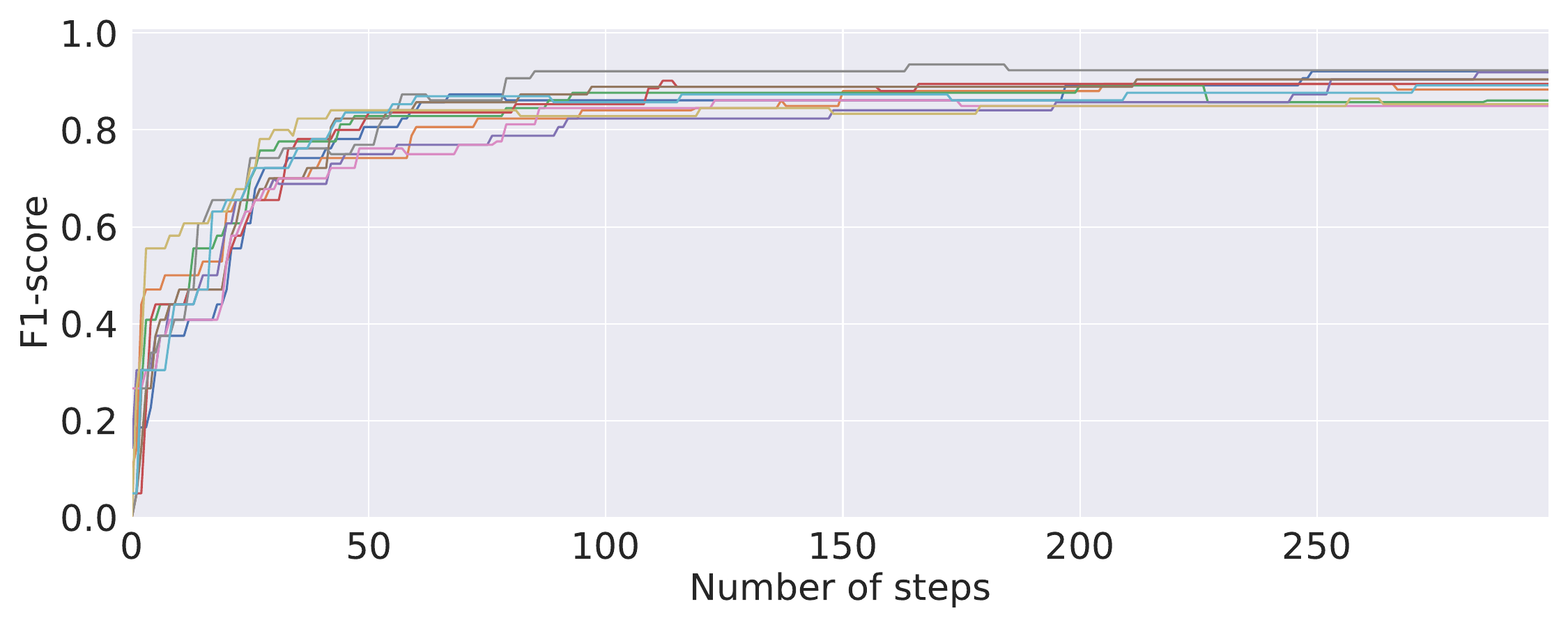}}
  \vspace{-.5em}
  \caption{Test set F1 as additional properties are learned and incorporated into the pool (10 framework runs).}
  \label{fig:f1}
\end{figure*}
\begin{figure*}[!ht]
  \centering
  \subfigure[Backblaze \(S1\).\label{fig:pool_smart}]{
    \includegraphics[width=0.32\linewidth,clip,trim={0 1em 0 1em}]{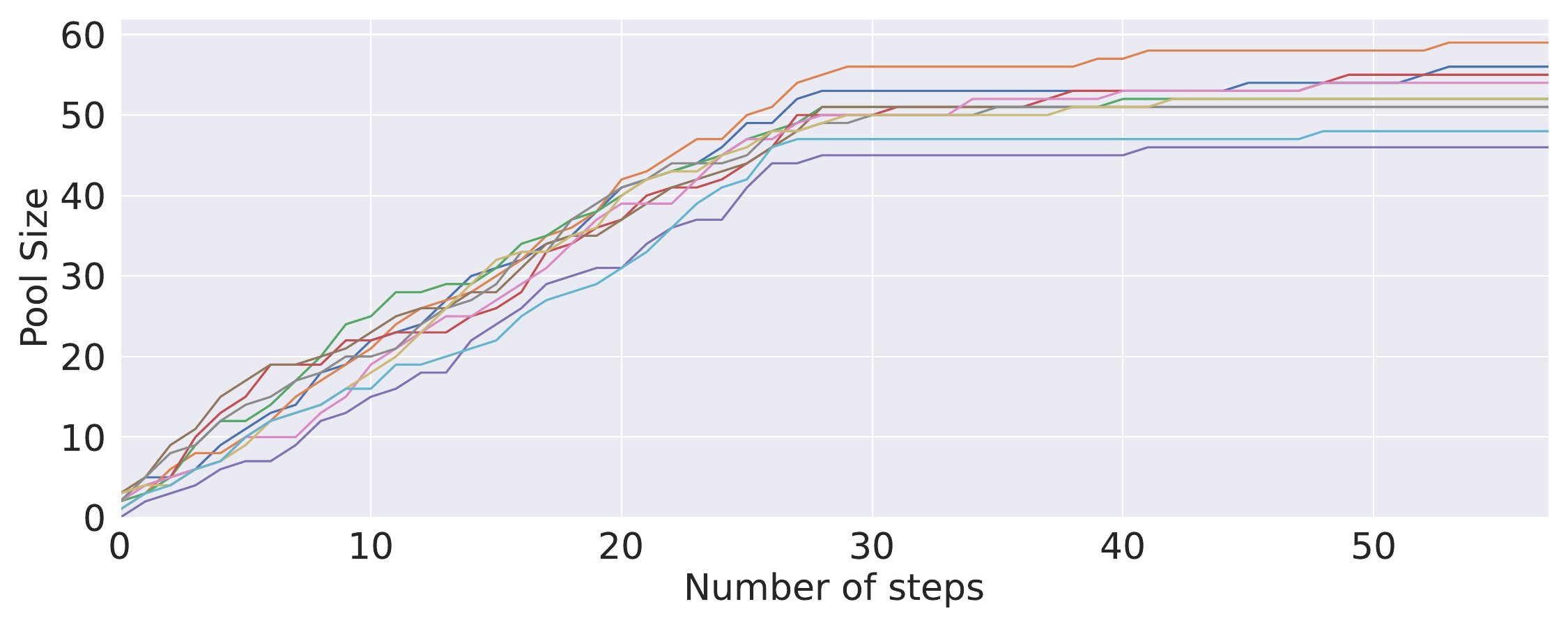}}
  \hfill
  \subfigure[TEP.\label{fig:pool_tep}]{
    \includegraphics[width=0.32\textwidth,clip,trim={0 1em 0 1em}]{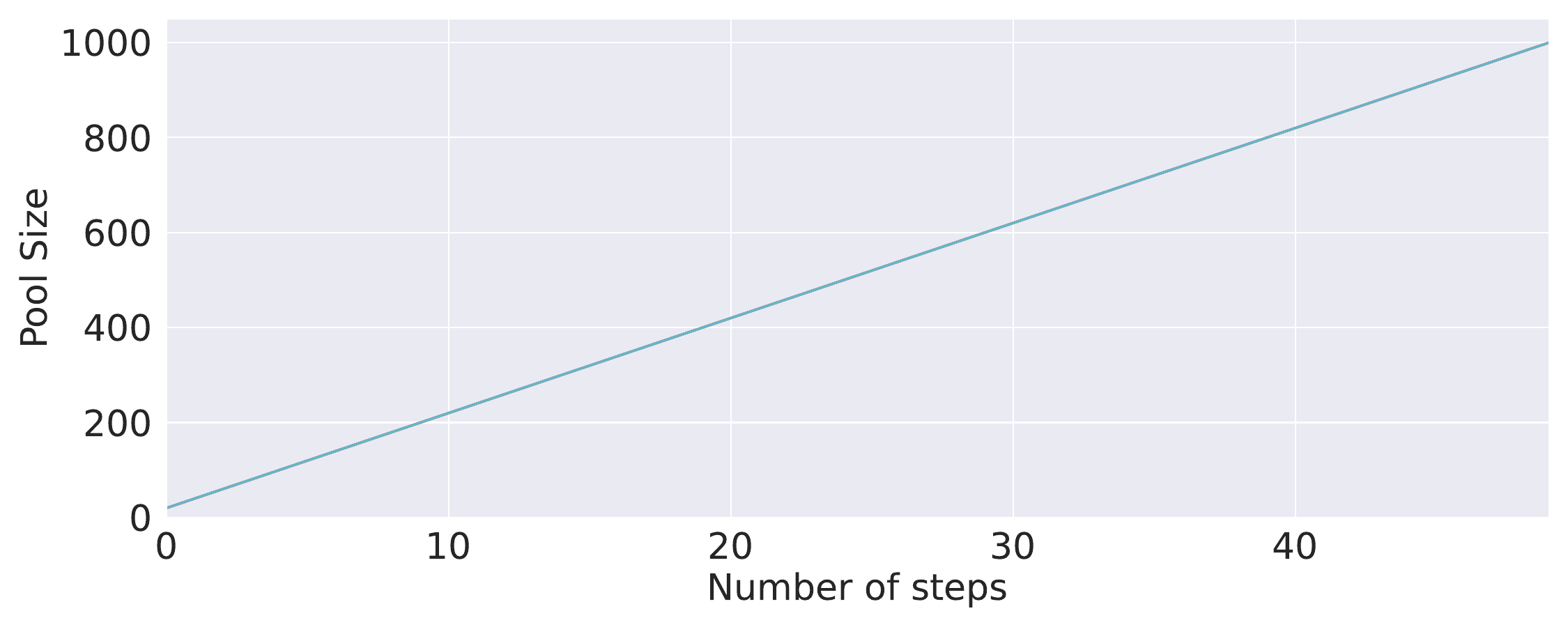}}
  \hfill
  \subfigure[C-MAPSS.\label{fig:pool_cmapss}]{
    \includegraphics[width=0.32\linewidth,clip,trim={0 1em 0 1em}]{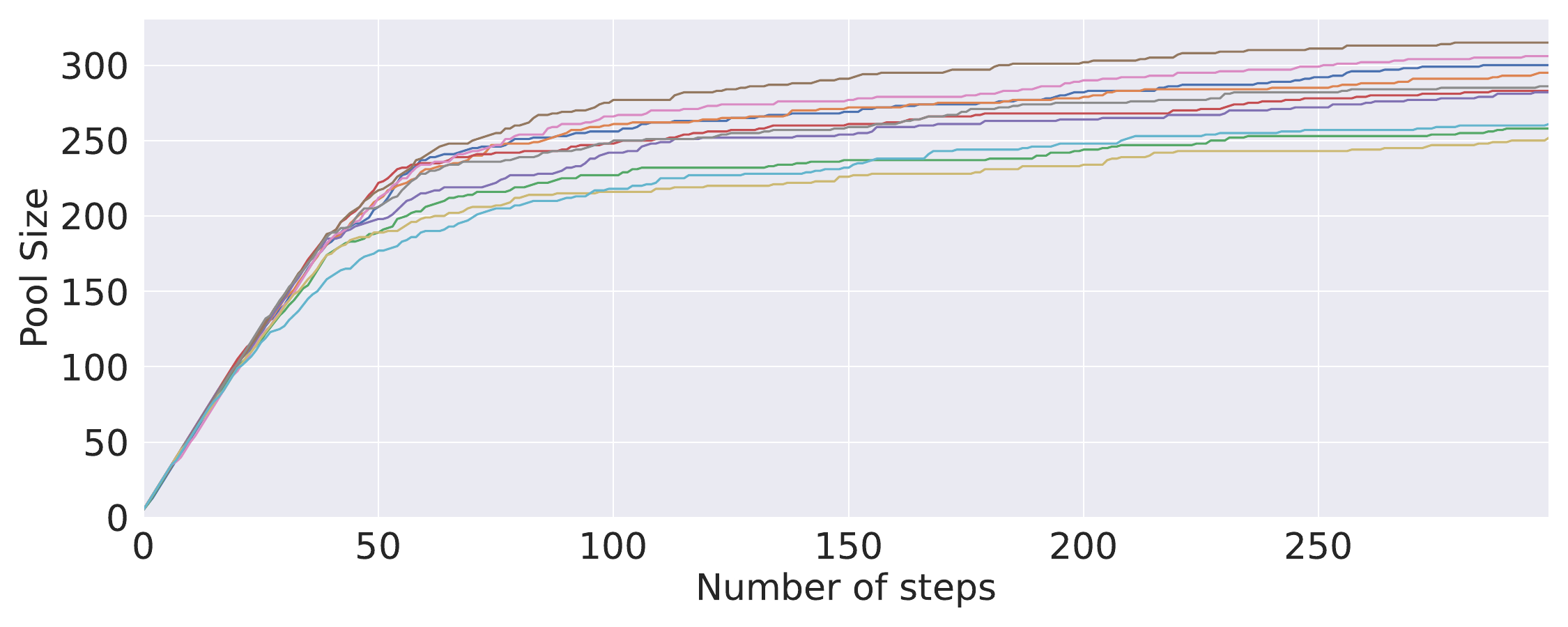}}
  \vspace{-.5em}
  \caption{Number of properties learned and added to the pool as training progresses (10 framework runs).}
  \label{fig:pool}
\end{figure*}

\Cref{tab:table_res} reports the main outcomes from the experimentation. %
Our approach outperforms the state-of-the-art in key metrics. In particular, it not only achieves the highest average F1 and MCC but also has a more stable behaviour over the experimental runs compared to %
GP-\GppSTL (best solutions of \cite{brunello2024learning}).\footnote{We statistically compared the 10 runs of GP-\GppSTL with those of our solution using a One-tailed ($H_1: \text{Our} - \text{GP-\GppSTL} > 0$) Wilcoxon Signed-Rank test ($\alpha = 0.05$), consistently finding a significant large difference %
for both F1 (Backblaze: p = .019, r = 0.7; C-MAPSS: p = .002, r = 0.9) and MCC (Backblaze: p = .007, r = 0.8; C-MAPSS:  p = .002, r = 0.9).}
In \cref{app:extended_results} and \cref{app:bench}, we respectively describe the metrics and a light benchmarking concretely showcasing the difference, in computing time, between using (CPU parallelized) \texttt{rtamt} and vectorized GPU-accelerated monitoring by trace checking. 
Also, given that our formula extraction method is based on heuristic search, in \cref{ap:convergence} we provide details about the EA's convergence behaviour.

Let us now consider how early the learned properties can identify a forthcoming failure. \cref{fig:anticip} illustrates the evolution of the average preemptiveness in failure detection on test set instances as additional formulas are learned over training data and incrementally added to the monitoring pool at each training step (similarly to batch updates during neural network training). 
In %
all the datasets, preemptiveness exhibits a pattern resembling logarithmic growth, characterized by a rapid initial increase followed by a period of stabilization.
A similar growth pattern can be observed for the F1 on the test set instances, as illustrated in \cref{fig:f1}, with the exception of TEP, which appears to reach performance saturation rather easily.
The increase in preemptiveness and F1 is closely tied to the number of formulas learned and incorporated into the monitoring pool as training progresses (\cref{fig:pool}). Once again, a logarithmic growth pattern is observed for Backblaze and C-MAPSS. Regarding TEP, note how no stabilization in the formula extraction is observed, leading to a large monitoring pool, despite the perfect F1 witnessed; a phenomenon that will be explored in future studies.

\subsection{Ablation study}

As previously mentioned, the main differences between our approach and the one described in \cite{brunello2024learning} are: 
\begin{inparaenum}[(i)]
    \item the introduction of a new bi-objective fitness function for the evolutionary algorithm; 
    \item the possibility of comparing signals within the formulas, not just a signal with a constant; 
    \item the incorporation of the concepts of epochs and batches in framework training;  
    \item the inclusion of a constants optimization step after the evolutionary phase of formula extraction.
\end{inparaenum}

\Cref{tab:ablation} details how much each of these modifications contributes to the performance improvements observed in our results, focusing on the dataset C-MAPSS. 
Using just the new fitness measure (1st row of the table), we achieve an average F1 of 0.81, representing a 2-point increase over the results reported in \cite{brunello2024learning}. 
Notably, as shown in the 2nd row of the table, introducing the capability to construct formulas where signals are compared to each other, rather than solely to constants, results in a slight performance decrease. This %
is not entirely surprising, as such comparisons increase the formula search space. On the other hand, the benefits of these comparisons are likely dataset-dependent, and they do not seem to benefit the C-MAPSS dataset, if not for a slightly higher preemptiveness. 
Next, incorporating the notions of epochs and batches during the framework’s training %
(3rd row of the table) yields a considerable performance improvement, in both F1 and preemptiveness. This %
can be attributed to the ability to iteratively learn from the training traces. 
The best performance is achieved when the constants optimization step is also integrated into the evolutionary algorithm (4th row of the table), thus obtaining our newly proposed full-fledged framework. This setup %
produces the best preemptiveness and the highest average F1. It also shows %
the greatest stability, %
having the lowest F1 standard deviation. 

\begin{table}[t]
\centering
\caption{Ablation study results, C-MAPSS test instances.\label{tab:ablation}}
\resizebox{\linewidth}{!}{%
\begin{threeparttable}
\begin{tabular}{lccccccc}
\toprule 
\textbf{Ablation case} & \textbf{P} & \textbf{R} & \textbf{FAR} & \multicolumn{3}{c}{\textbf{F1}} & \textbf{Prmt} \\
 & & & & \textbf{Min} & \textbf{Max} & \textbf{Avg\stdf{std}} \\
 \toprule
Only new fitness & .99 & .68 & .01 & .76 & .84 & .81\stdf{.03} & 14.32\\
Fit, multisignal & .95 & .68 & .02 & .76 & .85 & .79\stdf{.03} & 15.22 \\
Fit, multis, epochs & .94 & .87 & .03 & .84 & .96 & .89\stdf{.05} & 24.10 \\
Fit, multis, epochs, opt & .92 & .86 & .05 & .85 & .92 & .89\stdf{.03}  & 24.88 \\
\bottomrule
\end{tabular}
\begin{tablenotes}
\item Prmt = average failure preemptiveness, seconds (higher=better). P = precision; R = recall. 
\end{tablenotes}
\end{threeparttable}
}
\end{table}

\subsection{Formula interpretability analysis}
\label{sec:interpretability}

The good predictive performances of the properties learned by the framework are complemented by their inherent interpretability. 
Here, we present an example formula learned by the framework, focusing on the dataset Backblaze $S1$, since its domain is the most accessible; please refer to  \cref{ap:interpretability} for C-MAPSS and TEP. 
The formula is $\ltl{G}(power\_on\_hours < 28101.679897568785 \vee \ltl{H}(reported\_uncorrectable\_errors < 19.91357793883))$, which signals a failure (evaluating to $\bot$) if the following conditions are met: the hard drive has been in operation for an extended period of time (more than %
three years, assuming continuous use); and, the drive begins to show signs of degradation, when the number of reported errors that could not be resolved with error correction exceeds a threshold of around 19.91. %
The latter is indeed a critical metric to monitor, as emphasized in the Seagate's SMART documentation \cite{backblaze_smart_stats,backblaze_smart_stats_bis}.

\section{Conclusions and future work}
\label{sec:discussion_limitations}

We presented a GPU-enabled framework that leverages historical data to derive interpretable properties for early failure detection while providing formal guarantees. This is made possible by a seamless integration of monitoring and machine learning, supported by GPU acceleration and novel theoretical insights that reduce monitoring to trace checking, significantly speeding up formula learning. %
The proposed framework outperforms state-of-the-art approaches, while exhibiting a strong anticipatory behavior. 

We conclude by highlighting some future research directions to improve the framework.
First, \cref{alg:algorithm_training} learns formulas from a given supervised learning training set. 
We plan to investigate %
self-supervised methods to perform pseudo-labeling of traces. This represents a significant shift from failure detection to anomaly detection, where deviations from normal operation may not necessarily lead to system termination. This poses theoretical challenges, as the notion of monitoring in the context of anomalies is not yet well-established.
Second, the current algorithm does not account for the possible redundancy or obsolescence of formulas in the monitoring pool. Developing mechanisms to identify, update, or remove redundant or outdated formulas (potentially even after the initial training phase) is crucial for maintaining accuracy and efficiency over time.
Third, we will explore other methods to learn monitorable formulas. Currently, we rely on genetic programming; however, deep generative models may capture additional information, potentially leading to even better performance.
Finally, we acknowledge the need to support users who are not familiar with formal logic. In this respect, methodologies to translate natural language sentences into logical specifications, and vice versa, can significantly 
enhance usability and foster broader adoption of our framework.

\clearpage

\begin{ack}
    All the authors acknowledge the support from the 2024 Italian INdAM-GNCS
    project \lq\lq Certificazione, monitoraggio, ed interpretabilità in sistemi di intelligenza artificiale\rq\rq, ref. no. CUP E53C23001670001; and, the
    support from the Interconnected Nord-Est Innovation Ecosystem (iNEST),
    which received funding from the European Union Next-GenerationEU (PIANO NAZIONALE DI RIPRESA E RESILIENZA (PNRR) – MISSIONE 4 COMPONENTE 2,
    INVESTIMENTO 1.5 – D.D. 1058 23/06/2022, ECS00000043).
    N. Saccomanno acknowledges ISCRA for awarding the project SENTRY access to the LEONARDO supercomputer, owned by the EuroHPC Joint Undertaking, hosted by CINECA (Italy).
    A. Montanari acknowledges the support from the MUR PNRR project FAIR - Future AI Research
    (PE00000013) also funded by the European Union Next-GenerationEU.  This manuscript reflects only the authors’ views and opinions, neither the European Union nor the European Commission can be considered responsible for them.
\end{ack}

\bibliography{biblio_condensed}

\clearpage

\appendix

\section{Framework hyperparameters and tuning}
\label{app:hp_search}

\cref{tab:hyperpar} summarizes the framework’s hyperparameters along with a brief description. The tuning ranges and final values for each dataset are provided in \cref{tab:tuning}. 

\begin{table*}[t]
\centering
\caption{Hyperparameters of the framework.\label{tab:hyperpar}}
\resizebox{\linewidth}{!}{%
\begin{threeparttable}
\begin{tabular}{lll}
\toprule 
\textbf{Kind} & \textbf{Name} &\textbf{Description} \\
 \toprule
\multirow{3}{*}{{General}} 
 & $n_{aug\_fail}$ &   how many perturbed traces to generate for each failure trace \\
& $n_{aug\_good}$ &   how many perturbed traces to generate for each good trace \\
 & \minacc &   quality requirement for formula extraction: minimum accuracy over the splitting of augmented failure traces \\
 & \maxfar &   quality requirement for formula extraction: maximum FAR over augmented good traces \\
 & \batch &   batch size, i.e., number of pairs $(\sigma, A_\sigma)$ in each batch \\
 & \epoch &   number of epochs, i.e., number of iterations to perform over the training data \\
\midrule
\multirow{6}{*}{\begin{tabular}[c]{@{}l@{}}Genetic\\ algorithm\end{tabular}} 
& \fractgood &  relative size of sets $\Sigma_\bot'$, $\Sigma_\bot''$ with respect to the size of $\Sigma_\bot$\\
& \rotation &  rotation frequency, in number of generations, of $\Sigma_\bot'$  \\
& \maxgen &   maximum number of generations  \\
& \patience &   patience, in number of generations, of the early stopping criterion  \\
& \popsize &   number of individuals in the population  \\
& \mutprob &    mutation probability, which undergoes the following decay over the generations: $\mutprob / \sqrt[3]{generation\_index}$ \\
& \crossprob &   crossover probability \\
 \bottomrule
\end{tabular}%
\end{threeparttable}
}
\end{table*}

As described in the main paper, we conducted a tuning phase for each dataset. To such an extent, each training set was randomly partitioned into 90\% training and 10\% validation subsets, following a stratified approach based on the failure label. Additionally, system traces were grouped to prevent fragmentation between the two partitions.

For the hyperparameters in common with \cite{brunello2024learning}, we began with their recommendations, excluding the portions of the search space already discarded in their coarse evaluation phase and considering ranges of values close to those identified as optimal. 

We employed the \texttt{Hyperopt-sklearn} library\footnote{\url{https://github.com/hyperopt/hyperopt-sklearn}} with a budget of 50 iterations, since an exhaustive grid search over all the parameter combinations was computationally infeasible. The performance of each parameter combination was evaluated as the average F1 across three independent framework executions, with the objective of maximizing the F1.

Despite the numerous hyperparameters considered, note that some were assigned fixed values, while others consistently converged to the same tuned value across all datasets. These values can thus be regarded as sensible default choices. 

\begin{table}[t]
\centering
\caption{Hyper parameters tuning ranges and final values.\label{tab:tuning}}
\resizebox{\linewidth}{!}{%
\begin{tabular}{llllll}
\toprule 
\textbf{Kind} & \textbf{Name} & \textbf{Search range} & \textbf{Backblaze} & \textbf{C-MAPSS} & \textbf{TEP} \\
 \toprule
\multirow{3}{*}{{General}} 
 & $n_{aug\_fail}$ & [2, 5, 10, 20] & 5 & 20 & 2    \\
 & $n_{aug\_good}$ & [0, 2, 5, 10] & 0 & 10 & 2    \\
 & \minacc & [0.75]  & 0.75 & 0.75 & 0.75    \\
 & \maxfar & [0.0, 5e-5, 5e-4, 5e-3] & 5e-5 & 5e-3 &  0.0   \\
 & \batch & [5, 10, 20] & 5 & 5 & 20 \\
 & \epoch & [2, 5, 10, 15] & 2 & 15 & 2 \\
\midrule
\midrule
\multirow{6}{*}{\begin{tabular}[c]{@{}l@{}}Genetic\\ algorithm\end{tabular}} 
& \fractgood &  [0.33, 0.66] & 0.66 & 0.33 & 0.33  \\
& \rotation & [2, 5, 10] & 2 & 10 & 10    \\
& \maxgen & [500] & 500 & 500 & 500   \\
& \patience & [100] & 100 & 100 & 100   \\
& \popsize & [500] & 500 & 500 & 500   \\
& \mutprob & [0.3, 0.6] &  0.3  & 0.3  & 0.3    \\
& \crossprob & [0.7, 0.9] & 0.9 & 0.9 & 0.9     \\
 \bottomrule
\end{tabular}%
}
\end{table}

\section{Description of the metrics}
\label{app:extended_results}

We employed a comprehensive set of metrics to evaluate the framework's failure detection performance, including Precision, Recall, F1 Score, False Alarm Rate (FAR), and Matthews Correlation Coefficient (MCC). These metrics are derived from the confusion matrix, where positive instances correspond to failures and negative instances correspond to good traces. The confusion matrix is composed of the following elements: True Positives (TP), False Positives (FP), True Negatives (TN), and False Negatives (FN). 

Precision and Recall are particularly important in classification tasks like the one we considered. Precision measures the proportion of predicted failures that are actual failures and is defined as:

\begin{equation}
    \text{Precision} = \frac{TP}{TP + FP} \; .
\end{equation}

Recall quantifies the proportion of actual failures correctly identified and is crucial in scenarios where missing a failure can have significant consequences. It is defined as:

\begin{equation}
    \text{Recall} = \frac{TP}{TP + FN} \; .
\end{equation}

The F1 Score, a harmonic mean of Precision and Recall, is particularly useful when balancing the need to identify as many failures as possible (high recall) while ensuring the identified cases are indeed failures (high precision). It is calculated as:

\begin{equation}
    \text{F1 Score} = 2 \times \frac{\text{Precision} \times \text{Recall}}{\text{Precision} + \text{Recall}} \; .
\end{equation}

FAR, on the other hand, is critical in scenarios where false alarms carry a high cost. It measures the probability of falsely identifying a non-failure as a failure and is defined as:

\begin{equation}
    \text{FAR} = \frac{FP}{FP + TN} \; .
\end{equation}

Since FAR depends on TN, careful consideration is required in cases where good traces significantly outnumber failure traces, as is common in datasets with class imbalance.

Finally, the Matthews Correlation Coefficient (MCC) is defined as:

\begin{equation}
        \text{MCC} = \frac{(TP \times TN) - (FP \times FN)}{\sqrt{(TP + FP)(TP + FN)(TN + FP)(TN + FN)}} \; .
\end{equation}

The MCC ranges from -1 to +1, where +1 signifies perfect prediction, 0 indicates no better performance than random guessing, and -1 represents total disagreement between predictions and actual outcomes. By incorporating all four categories of the confusion matrix (true positives, true negatives, false positives, and false negatives), MCC serves as a robust and balanced metric for evaluating model performance. 
We chose to include MCC in our analyses due to its effectiveness in handling class imbalance. This provides a more comprehensive assessment of the framework’s predictive performance, addressing limitations in other metrics that can be misleading in the presence of rare failure events \cite{chicco2020advantages}.

\section{Benchmarking RTAMT and Parser}
\label{app:bench}

For the evaluations, we utilized a Dell PowerEdge R750 server running Red Hat Enterprise Linux 8.7 (Ootpa). The system is equipped with two Intel(R) Xeon(R) Platinum 8360Y CPUs operating at 2.40 GHz, providing a total of 72 physical cores. The machine also features 512 GB of memory, 4 TB of disk space, and an NVIDIA A100 GPU with 80 GB of VRAM. 

We compared the performance, in terms of computation time, of monitoring and trace checking. For each experiment we conducted three executions, setting a cumulative running time limit of 180 seconds. Computations were terminated if the maximum allowed running time was exceeded or if available system or GPU memory was insufficient. 
For monitoring, we considered RTAMT when running on a single core, and when parallelized (with up to 128 virtual cores) either over the formulas or over the traces. 
For trace checking, we implemented it by means of a Lark parser \cite{lark_parser}, running vectorized computations across both traces and formulas over the GPU. 

\begin{figure}[t]
  \centering
  \subfigure[Varying the trace length.\label{fig:bench_length}]{
    \includegraphics[width=\linewidth,clip]{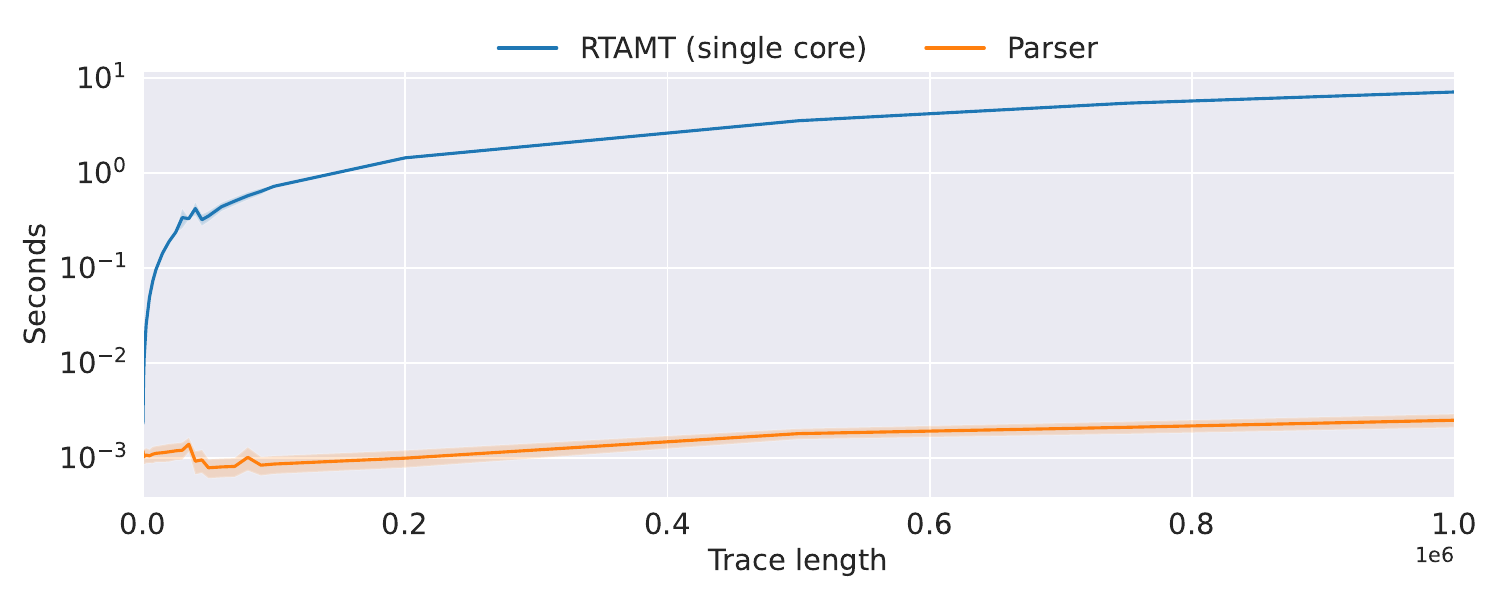}}
  \vspace{-0.25em}

  \subfigure[Varying the number of traces.\label{fig:bench_traces}]{
    \includegraphics[width=\linewidth,clip]{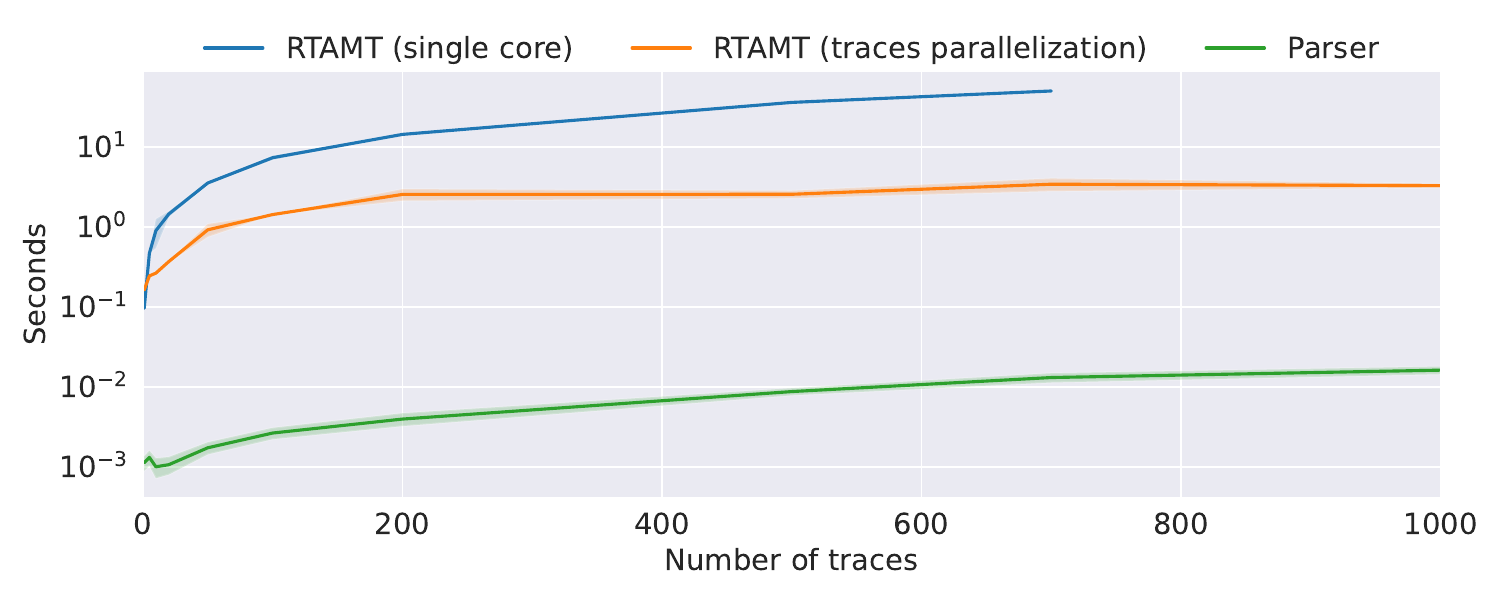}}
  \vspace{-0.25em}

  \subfigure[Varying the number of formulas.\label{fig:bench_formulas}]{
    \includegraphics[width=\linewidth,clip]{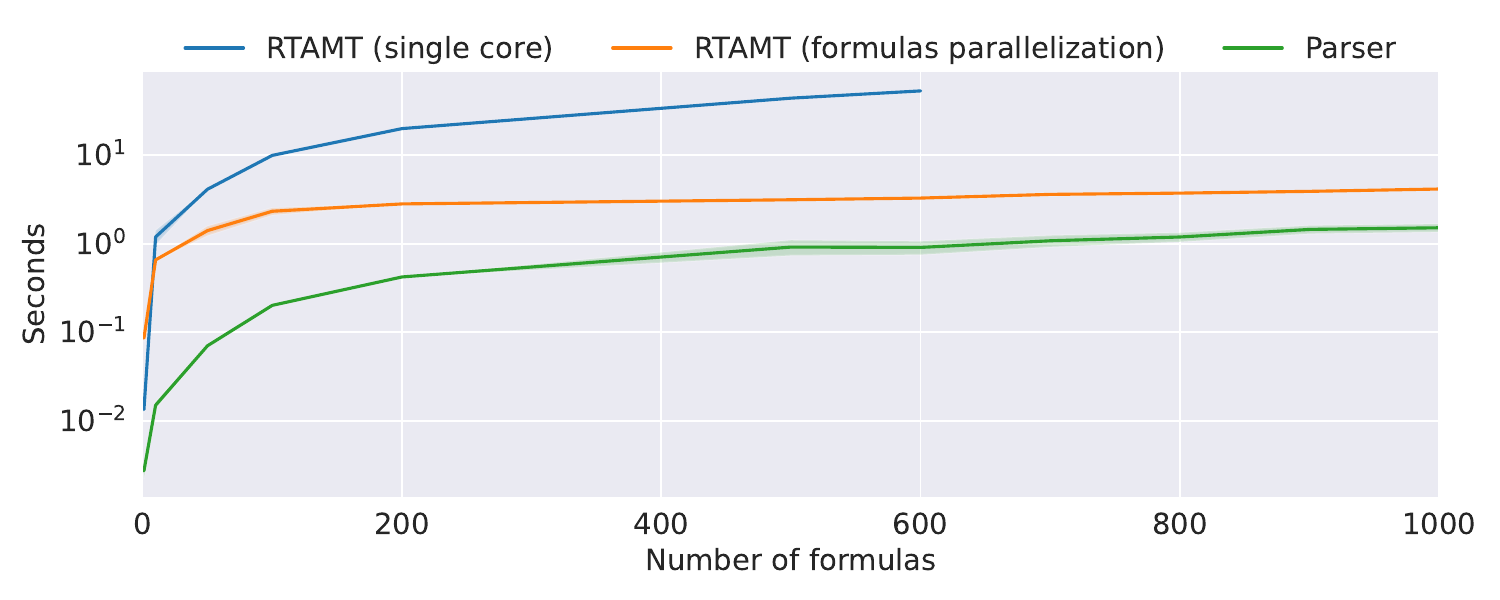}}

  \caption{Comparison of the computation time (log scale) between monitoring (RTAMT) and trace checking (Lark). Solid lines show the average runtimes; shaded areas indicate standard deviations.}
  \label{fig:bench}
\end{figure}

To begin with, we drew inspiration from \cite{yamaguchi2024rtamt}, which investigates the scalability of RTAMT by monitoring selected properties with respect to a single random trace while varying its length. Here, we consider the property $\ltl{O}(x_1 >= 0.3) \xrightarrow{} \ltl{H}(x_2 >= 0.1)$, that includes both temporal and classical boolean operators. \cref{fig:bench_length} reports the results. The runtime of RTAMT monitoring (single core) is closely aligned with that of the Python implementation discussed in \cite{yamaguchi2024rtamt}, and is overall three to four orders of magnitude slower than trace checking using the Lark parser.

We then extended the analysis to evaluate scalability in terms of the number of random traces, while fixing each trace length at 10,000. As shown in \cref{fig:bench_traces}, trace checking once again significantly outperforms monitoring, even when RTAMT is parallelized across traces. Notably, the single core implementation of RTAMT encounters a timeout when the number of traces exceeds 700.

Finally, we evaluated scalability with respect to the number of formulas. For this analysis, we used a single bi-variate trace of length 1,000. Formulas, of depth 5, were randomly generated following the \ppSTL syntax. Results are shown in \cref{fig:bench_formulas}. RTAMT times out after 600 formulas when run without parallelization, which was to be expected given its runtime over a single trace shown in \cref{fig:bench_length}; however, parallelization allows it to handle up to 1,000 formulas. Trace checking still delivers the fastest computation, though here the speed improvement is smaller than when scaling over the trace length or the number of traces. This can be explained by the fact that the trace checking benchmarking includes not only the time spent on calculating the robustness but also the additional time required to create the concatenated string representation of the formulas and to parse and translate it into a Python partial formula.

\begin{table*}[t]  
\centering
\caption{Reference symbol table.}\label{tab:symbols}
\begin{tabularx}{\textwidth}{@{}>{\raggedright\arraybackslash}p{0.2\textwidth} X@{}}
\toprule
\textbf{Symbol} & \textbf{Meaning / Definition} \\
\midrule
\(\sigma\) & Trace: finite, nonempty sequence of states. Domain \((\mathbb{R}^n)^+\). \\
\(|\sigma|\) & Length (number of states) of trace \(\sigma\). \\
\(\sigma_{[0,i]}\) & Prefix of \(\sigma\) up to position \(i\). \\
\(\lang\) & A language: a set of traces. \\
\(\rho(\varphi,\sigma,i)\) & Robustness (quantitative semantics) of formula \(\varphi\) on trace \(\sigma\) at position \(i\). \\ 
\(\sigma,i\models\varphi\) & Qualitative satisfaction iff \(\rho(\varphi,\sigma,i)\ge 0\). \\
\(\sigma\models\varphi\) & Models: satisfaction at first (resp., last) position for $\mathsf{STL}$ (resp., $\mathsf{ppSTL}$). \\
\(\lang(\varphi)\) & Language of \(\varphi\): \(\{\sigma\in(\mathbb{R}^n)^+\mid \sigma\models\varphi\}\). \\ 
\(\mathsf{G}(\mathsf{ppSTL}),\;\mathsf{F}(\mathrm{\mathsf{ppSTL}})\) & Safety / cosafety syntactic fragments: \(\mathsf{G}(\psi)\) or \(\mathsf{F}(\psi)\) with \(\psi\in\mathsf{ppSTL}\). \\ 
\(\mathrm{mon}_{\lang}\) & Monitor for language \(\lang\): \((\mathbb{R}^n)^+\to\{\top,\bot,?\}\): \(\top\) iff all continuations stay in \(\lang\); \(\bot\) iff all continuations stay out; else \(?\).\\
\(\mathrm{mon}_\varphi\) & \(\mathrm{mon}_\varphi:=\mathrm{mon}_{\lang(\varphi)}\) . \\
\(T\subseteq(\mathbb{R}^n)^+,\;m,\;l\) & Set of traces, \(m=|T|\), \(l=\max_{\tau\in T}|\tau|\). Represented as tensor \(T\in\mathbb{R}^{m\times l\times n}\) (padding as needed). \\
\(g_\varphi\) & Vectorized robustness function of formula $\varphi$, $g_\varphi : \mathbb{R}^{m \times l \times n} \to \mathbb{R}^{r \times m \times l }$. \\
\({rob}_{\mathbf{T},\varphi}\) & Robustness matrix with \({rob}_{\mathbf{T},\varphi}[k,j]=\rho(\varphi,T[k],j)\). \\
\(\Sigma_\top,\;\Sigma_\bot\) & From framework input dataset \(\mathcal{X}\): sets of failure traces (\(\top\)) and good traces (\(\bot\)). \\
\(A_\sigma\) & Augmented traces generated from a failure trace \(\sigma\). \\
$\mathcal{P}$ & Within the framework, pool of formulas. \\
$\phi_\sigma$ & Within the framework, formula $\phi$ linked to a set $(\sigma, A_\sigma)$.\\
\({rob}_{t,\phi_\sigma},\ \max\mathrm{Rob}(t,\phi_\sigma)\) & Robustness vector for formula $\phi_\sigma$ over trace \(t\); and its maximum over time points. \\
\({score}_{t,\phi_\sigma}[i]\) & \lq\lq Goodness\rq\rq{} of the split point $i$ of trace $t$, according to formula $\phi_\sigma$ (${score}_{t,\phi_\sigma}$ is a vector). \\
\(\mathrm{margin}(\phi_\sigma)\) & Worst-case best split score over \(\{\,\sigma\,\}\cup A_\sigma\). \\
\(\mathrm{okOrig}(\phi_\sigma)\) & Checks \(\mathrm{rob}_{\sigma,\phi_\sigma}[0]<0\) and \(\max\mathrm{Rob}(\sigma,\phi_\sigma)\ge0\). \\
\(\mathrm{acc}(\phi_\sigma)\) & Fraction of \(\{\,\sigma\,\}\cup A_\sigma\) where a good-prefix/bad-suffix split is achieved. \\ 
\(\mathrm{far}(\phi_\sigma)\) & False alarm rate over \(\Sigma_\bot\): share of good traces with \(\max\mathrm{Rob}\ge0\). \\
\(\mathrm{goodRob}(\phi_\sigma,S)\) & Worst-case max robustness on a sampled good set \(S\in\{\Sigma'_\bot,\Sigma''_\bot\}\). \\
\bottomrule
\end{tabularx}
\end{table*}

\section{Convergence of the EA}
\label{ap:convergence}

Our framework relies on a EA-based approach for formula extraction. As with any other heuristic search method, it is important to evaluate the convergence behaviour of the EA. 
Recall that the evolutionary process for formula extraction is executed within our framework (see \cref{alg:algorithm_training}) multiple times per dataset with different random seeds. In particular, given a seed, in dataset C-MAPSS it is run 300 times, in dataset Backblaze 58 times and in dataset TEP 50 times. 
Each run has a budget of 500 generations (\cref{tab:tuning}) and employs early stopping based on the hypervolume metric: if the hypervolume does not improve for 100 consecutive generations (\cref{tab:tuning}), the run is terminated for convergence. Early stopping is triggered in $\sim$85\% of runs on C-MAPSS, $\sim$100\% on Backblaze, and $\sim$78\% on TEP (which nevertheless reaches a perfect score), indicating that the optimization budget is adequate and that the EA indeed converges.

\section{Extended formula interpretability analysis}
\label{ap:interpretability}

Here, we present examples of formulas learned and added to the framework’s monitoring pool during its training phase on C-MAPSS and TEP datasets. As can be seen, they are highly interpretable. Recall, as discussed in the main article, that they represent safety properties belonging to $\GppSTL$, thus encoding always-desired good behaviors of the system. As the framework’s training progresses, the violation of these formulas enables to identify undesirable scenarios earlier and earlier, effectively capturing the gradual progression of failure as it unfolds.

\paragraph{C-MAPSS}
Consider the following chain of formulas, that signal a forthcoming failure when they evaluate to $\bot$. At a certain point, during the framework’s training process, the formula %
$\ltl{G}(corrected\_core\_speed_{rpm} < 8176.38427312112 \vee LPC\_outlet\_temperature_{R} < 643.5223271161558)$ is added into the monitoring pool. Subsequently, to enable earlier failure detection, the formula %
$\ltl{G}(corrected\_core\_speed_{rpm} < 8183.030687071466)$
is included. Further tracing back in time along the failure trajectories, the framework identifies the formula %
$\ltl{G}(\ltl{H}[6,20](corrected\_core\_speed_{rpm} < 8175.1750026268))$, which should capture even earlier indications of system degradation. Note how all these formulas revolve around tracking the value of $corrected\_core\_speed_{rpm}$. This aligns with findings from other studies conducted on the dataset, which highlight how high values of such an attribute may contribute to a forthcoming failure (see, e.g., \cite{alomari2024shap,youness2023explainable}).

\paragraph{TEP}
This formula was added to the monitoring pool: $\ltl{G}(\ltl{O}[187,188](\ltl{O}[22,180] (D\_feed\_flow\_valve \geq stripper\_underflow)) \wedge composition\_of\_C\_purge < 23.887858867109713)$. 
In short, to avoid a failure, at any time the formula requires a maximum purge composition related to the stream feed \lq\lq C\rq\rq{}, and it also requires evidence in the past that, at least once, the value of the flow valve of stream feed \lq\lq D\rq\rq{} was greater or equal than the value of $Stripper\_underflow$. 
While a complete understanding of the formula requires non-trivial domain knowledge (see, e.g., \cite{tepapp}), a possible interpretation is as follows. By monitoring how much \lq\lq C\rq\rq{} appears in the purge (at most around 23.89), it is possible to ensure safe operation. Simultaneously, checking whether the \lq\lq D\rq\rq{} feed-flow valve has historically been open sufficiently ($\geq$ stripper underflow) is a way of ensuring that enough of component \lq\lq D\rq\rq{} was introduced to keep the reactor balanced, or possibly that the stripper was not starved. Indeed, in the scenario of TEP, maintaining the correct ratio of inlet feeds versus outflows is important to keep the reactor balanced and operationally safe.

\section{Data and code availability}
\label{app:reprod}

We provide all the necessary code to allow interested researchers to reproduce our experiments and apply our framework to their domains for novel applications.

\smallskip

\texttt{LINK:} \pdfulink{https://github.com/dslab-uniud/interpretable-failure-detection-gpu}

\smallskip

The provided materials %
include the code for learning a pool of formulas using our framework, dictionaries containing the hyperparameters obtained from the tuning phase, the three datasets utilized in our study, a list of Python packages (with their respective versions) required to reproduce the experiments, and the code for testing the results of a framework execution. 
For detailed instructions on executing the code and reproducing the experiments, please refer to the \texttt{readme.md} file located in the home folder.

\end{document}